\documentclass[preprint]{elsarticle}
\usepackage{adjustbox}
\usepackage{subcaption}
\usepackage{xcolor}
\usepackage{lineno,hyperref}
\usepackage[ruled,vlined]{algorithm2e}
\usepackage{dingbat}
\modulolinenumbers[5]

\journal{Computer Vision and Image Understanding}
\usepackage{etoolbox}
\makeatletter
\patchcmd{\ps@pprintTitle}
  {Preprint submitted to}
  {Published in}
  {}{}
\makeatother

\begin{document}

\begin{frontmatter}

\title{A Multi Camera Unsupervised Domain Adaptation Pipeline for Object Detection in Cultural Sites through Adversarial Learning and Self-Training}

\author[1]{Giovanni Pasqualino} 
\author[1]{Antonino Furnari\corref{cor1}}
\cortext[cor1]{Corresponding author:}
\ead{furnari@dmi.unict.it}
\author[1,2,3]{Giovanni Maria Farinella}

\address[1]{Department of Mathematics and Computer Science, University of Catania, Viale Andrea Doria 6, Catania, 95125, Italy}
\address[2]{CUTGANA, University of Catania, Via Santa Sofia 98, Catania, 95123, Italy}
\address[3]{ICAR-CNR, National Research Council, Via Ugo la Malfa 153, Palermo, 90146, Italy}

\begin{abstract}
Object detection algorithms allow to enable many interesting applications which can be implemented in different devices, such as smartphones and wearable devices. In the context of a cultural site, implementing these algorithms in a wearable device, such as a pair of smart glasses, allow to enable the use of augmented reality (AR) to show extra information about the artworks and enrich the visitors' experience during their tour. However, object detection algorithms require to be trained on many well annotated examples to achieve reasonable results. This brings a major limitation since the annotation process requires human supervision which makes it expensive in terms of time and costs. A possible solution to reduce these costs consist in exploiting tools to automatically generate synthetic labeled images from a 3D model of the site. However, models trained with synthetic data do not generalize on real images acquired in the target scenario in which they are supposed to be used. Furthermore, object detectors should be able to work with different wearable devices or different mobile devices, which makes generalization even harder. In this paper, we present a new dataset collected in a cultural site to study the problem of domain adaptation for object detection in the presence of multiple unlabeled target domains corresponding to different cameras and a labeled source domain obtained considering synthetic images for training purposes. We present a new domain adaptation method which outperforms current state-of-the-art approaches combining the benefits of aligning the domains at the feature and pixel level with a self-training process. We release the dataset at the following link \url{https://iplab.dmi.unict.it/OBJ-MDA/} and the code of the proposed architecture at \url{https://github.com/fpv-iplab/STMDA-RetinaNet}.
\end{abstract}


\end{frontmatter}

\linenumbers
\nolinenumbers
\section{Introduction}
\begin{figure}[t!]
\includegraphics[width=1\textwidth]{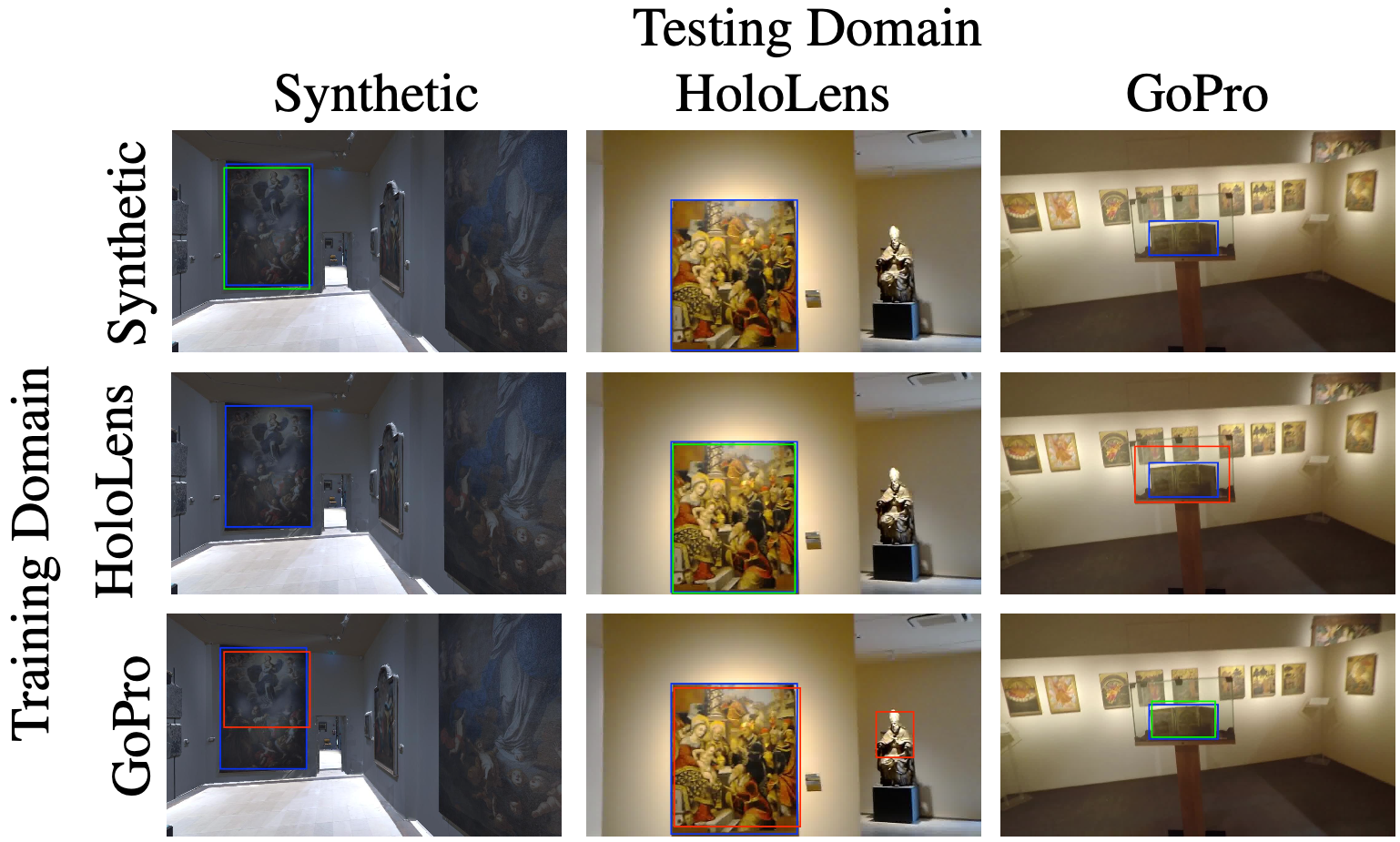}

\caption{Qualitative results of a standard object detector trained and tested on different domains. Blue bounding boxes represent the ground truth, green boxes represent correct detections, whereas red boxes indicate wrong detections (either object localization or classification). The model was trained using the domain indicated in the rows and tested using the domain reported in the columns.}
\label{fig:resultnoadapt}
\end{figure}

In recent years, wearable and mobile devices have increasingly attracted the interest of the scientific community because of their ability to capture human-centric data which reflects the intent and interests of the users (e.g., a picture shot with a mobile phone or a video acquired with an action camera). Given their ever-increasing computing capabilities, different computer vision algorithms have been integrated into these devices allowing the development of new applications in different scenarios.
In particular, previous works, have shown that human-centric devices such as mixed reality glasses can be exploited in a cultural site to improve the fruition of artworks~\cite{10.1145/3092832,articlecucchiara} by showing extra information to visitors using augmented reality (AR), or to track users' behavior~\cite{vedi2019}. While all these applications require the ability to detect objects in the scene, training object detectors is still costly and time consuming because it requires images to be labeled by human annotators. To reduce the data annotation costs, the authors of~\cite{orlando2020egocentric} proposed to create large datasets of images of a cultural sites by generating synthetic labeled images in a simulated environment. Despite this approach speeds up data collection and reduces the annotation costs, object detectors trained with synthetic images achieve poor performance when tested with real images due to the domain shift between the data used for training (source domain) and the data used at test time (target domain)~\cite{chen2018domain}. Furthermore in real-workly scenarios, object detection algorithms often need to be deployed to different devices, which are generally equipped with different cameras. This constraint further reduces the generalization ability of the object detection methods in real scenarios. Figure~\ref{fig:resultnoadapt} reports some qualitative results of a standard object detector trained and tested on different domains of images of a cultural sites: a set of synthetic images, real images acquired with an HoloLens device, and a set of images collected with a GoPro. As can be noted, the detection of the artworks works perfectly only if the the training and test set belong to the same data distribution.\newline
Domain adaptation techniques~\cite{ganin2014unsupervised} can be used to reduce the domain difference between source and target sets. However, in a real scenario, the algorithm should also generalize to images collected using multiple cameras as in the example in Figure~\ref{fig:resultnoadapt}, which may present subtle characteristics capable of affecting model performance.
We propose to tackle this problem as a multi-target unsupervised domain adaptation task in which there is a labeled source domain (the synthetic data) and more than one unlabeled target domains (the target images acquired using different cameras). We note that, since target unlabeled images can be acquired with a little effort, the task setup involves a small additional overhead as compared to single-target domain adaptation. We hence investigate whether the presence of more than one target domains can assist the domain adaptation process in the considered settings. To analyze the problem, we introduce a new dataset of both synthetic and real images collected in a cultural site and suitable to study unsupervised multi-camera domain adaptation. We perform experiments to assess the ability of current domain adaptation approaches to generalize across multiple cameras. We hence investigate a generalization of current state-of-the-art methods which is shown to outperform current methods. The contributions of this paper are as follows. 1) We present the first dataset to study multi-camera domain adaptation in cultural sites. The dataset has been acquired by real visitors in a cultural site using two different wearable cameras. 2) We propose a domain adaptation approach which takes advantage of multiple unlabeled camera domains and a self-training procedure to improve cross-domain generalization. The proposed method outperforms the results of current state-of-the-art methods by up to +23\% mAP. We discuss the limit of the proposed technique and present possible future research directions. The reminder of this paper is organized as follows. In Section~\ref{related} we discuss related work. Section~\ref{dataset} presents the proposed dataset. Section~\ref{methods} discusses the proposed pipeline and method. Section~\ref{results} reports the experimental settings and discusses results. Section~\ref{conclusion} concludes the paper and summarises the main findings of our study.

\section{Related Work}
\label{related}
In this section, we discuss the lines of research related to this work: use of wearable devices in cultural site, unsupervised domain adaptation techniques and unsupervised domain adaptation approaches for object detection.
\subsection{Use of wearable device in cultural site}
Wearable device applications allow to improve the fruition and the perception of the reality around us. In the context of cultural sites, the authors of~\cite{10.1145/3092832,portaz} focused on the creation of virtual guides to enrich the visitors’ experience with the fruition of multimedia materials. The applications are based on the detection and recognition of objects to trigger the presentation of associated information. Training these kinds of algorithms requires many labeled images that must be acquired and manually annotated, thus increasing development times and costs. Due to the lack of data in this field, the authors of~\cite{Ragusa_2020} proposed a dataset of first person videos acquired using Microsoft HoloLens to study different problems in the context of cultural sites.  
The authors of~\cite{orlando2020egocentric} presented a tool to generate synthetic labeled images from a 3D reconstruction of a cultural site. However, the generated images differ in color and shape with respect to the real counterparts. For this reason, object detection algorithms, which have to work at inference time on real images, produce poor result if trained only with synthetic data.\newline
In this paper, we study the problem of object detection in the presence of multiple target domains acquired using different wearable cameras. We present an approach to train an object detector to maximize its detection and recognition accuracy on both target camera domains using labeled synthetic images and unlabeled real images captured with the two cameras.

\subsection{Unsupervised Domain Adaptation}
Unsupervised domain adaptation (UDA) algorithms transfer the knowledge learned from a source labeled domain to a target unlabeled domain.
Most state-of-art domain adaptation approaches take into account labeled source and an unlabeled target domains. Past works in this field focused on reducing some divergence statistics between representation of example belonging to the two domains. The authors of~~\cite{Rozantsev_2019} proposed to use the MMD~\cite{gretton2008kernel} metric to reduce the distributions difference between source and target feature distributions. The authors of~~\cite{Sun_2017} proposed the CORAL metric and integrated it inside a CNN to align the covariances of the source and target feature distributions. 
Other methods propose solutions based on adversarial training. The authors of~\cite{ganin2014unsupervised} presented the Gradient Reversal Layer (GRL) which mimics the behavior of a generative adversarial network (GAN)~\cite{goodfellow2014generative} to encourage the extraction of indistinguishable features from images belonging to the source and target domains. The gradient reversal layer allows to train the architecture end-to-end instead of using the usual alternate optimization of the generator and the discriminator as in GANs~\cite{goodfellow2014generative}. The authors of~\cite{Tzeng_2017} proposed a method based on two stages that combines discriminative modeling, untied weight sharing, and a adversarial loss. The authors of~\cite{CycleGAN2017} presented a method based on image to image translation that, in the absence of paired example, learns a mapping between the two domains through the use of the adversarial loss. The authors of~\cite{du2021crossdomain} proposed a clustering based method to generate pseudo labels for the target domain, than the method minimizes the discrepancy of the gradients generated by the source and target images.
Other works studied the unsupervised domain adaptation problem in the presence of many labeled source domains and only one unlabeled target domain. The authors of~\cite{zhao2018adversarial} proposed a method based on the gradient reversal layer that discriminates between all of $Target-Source_n$ pairs where $d = 0,...,D$ and $D$ is the number of source domains. The authors of~\cite{peng2019moment} presented a method that consists of three components: feature extractor, moment matching module and a final classifiers.
These methods require the access to multiple labeled source dataset which in some cases maybe available or easy to produce.
Another line of research focused on a more realistic scenario where there is only one source domain and multiple target domains. In this case, the presence of multiple target domains emulate a real scenario where, for example, different devices with different lenses and images generation pipelines (IGP) produce different target domain. The authors of~\cite{gholami2020unsupervised} proposed a method based on an autoencoder which finds a latent space which can capture domain invariant and domain dependent features that can generalize over multiple target domains. The authors of~\cite{ragab2020adversarial} presented a method that extends the idea proposed by \cite{Tzeng_2017} replacing a binary discrimination with a multi-class discrimination. The authors of~\cite{nguyen2021unsupervised} proposed a method based on an iterative multi-teacher knowledge distillation from multiple teachers to a common student.\newline
The presented methods which tackle the multi target domain adaptation problem are used to solve classification task and they are not directly applicable to the object detection problem. In this work, we study how to exploit the information coming from the different target domains for the object detection adapting an adversarial training scheme similar to the work of the authors of~\cite{ganin2014unsupervised}.

\subsection{Unsupervised Domain Adaptation for Object Detection}
The methods described in the previous section can be adapted to consider the object detection task. The authors of~\cite{chen2018domain} presented DA-Faster RCNN  which is a modified version of Faster RCNN~\cite{DBLP:journals/corr/RenHG015} which aligns features at the image and instance levels exploiting gradient reversal layers~\cite{ganin2014unsupervised}. The authors of~\cite{Saito_2019} proposed to adapt the high-and low-level features. \cite{Xie_2019} proposed to extend the architecture presented by the authors of~\cite{chen2018domain} adding more discriminators with gradient reversal layers to the Faster RCNN backbone.
The authors of~\cite{kim2019diversify} presented an approach composed of two stages: 1) a domain diversification stage where the distribution of the labeled data is diversified by generating various distinctive domains shifted from the source domain using image to image techniques; 2) multi-domain-invariant representation learning, where adversarial learning is applied with a multi-domain discriminator to encourage feature to be indistinguishable across domains.
The authors of~\cite{yu2021unsupervised} proposed to translate images from the source domain into the target domain using CycleGAN and trained an object detector using a self-training procedure to create pseudo label for the target dataset. The authors of~\cite{9010241} introduced in a SSD architecture a novel self-training method called weak self-training (WST) combined with the adversarial background score regularization (BSR) to prevent the degeneration of the performance due to incorrect pseudo label obtained using a naive approach reducing the amount of false negative and positive detections. The authors of~\cite{9093358} presented a framework which combines intermediate domains to progressively adapt feature alignment for object detection and a weighted task loss which weights the samples in the intermediate domain. The authors of~\cite{FUJII2021100071} presented a method based on SSD which is divided in three steps: in the first step the SSD detector is pretrained using the source images; in the second step the source images are converted to real with CycleGAN; in the third step SSD is trained using the converted source images, the target images and using the weak self-training method proposed in~\cite{9010241}. The authors of~\cite{chen2021i3net} proposed a Implicit Instance-Invariant Network ($I^3Net$), a single stage object detector which adapt the source and the target domain considering: 1) a strategy to assign large weights to those sample-scarce categories and easy-to-adapt samples considering the intra-class and intra-domain variation, 2) a module to suppress uninformative background features boosting the foreground object matching, 3) a module that align the category at different domain specific layers and regularize the average prediction of different layer respect to the same category. The authors of~\cite{vidit2021attentionbased} introduced a generic approach based on an attention mechanism which allows to detect the important regions of the feature map extracted from the backbone on which adaptation should focus.
The authors of~\cite{guan2021uncertaintyaware} proposed a method which works at image and instance level aligning the two distributions so that well-aligned and poor-aligned samples are adaptively weighted based on the uncertainty of each sample. The authors of~\cite{vs2021megacda} presented a feature alignment method based on Faster RCNN which consist of three modules: 1) a global discriminator which align the feature extracted from the backbone; 2) category wise discriminators which aligns the features of each class belonging to the source and the target domains; 3) a memory guided attention mechanism which aids the category-wise discriminators to align category specific features between the two domains.\newline
Our work investigates whether the presence of multiple unlabeled target domains can improve the generalization of current methods. We further present an architecture based on feature alignment, image to image translation and self-training to tackle multi-camera unsupervised domain adaptation for object detection in cultural sites.
\begin{figure*}[t!]
            \centering
            \includegraphics[width=.12\textwidth]{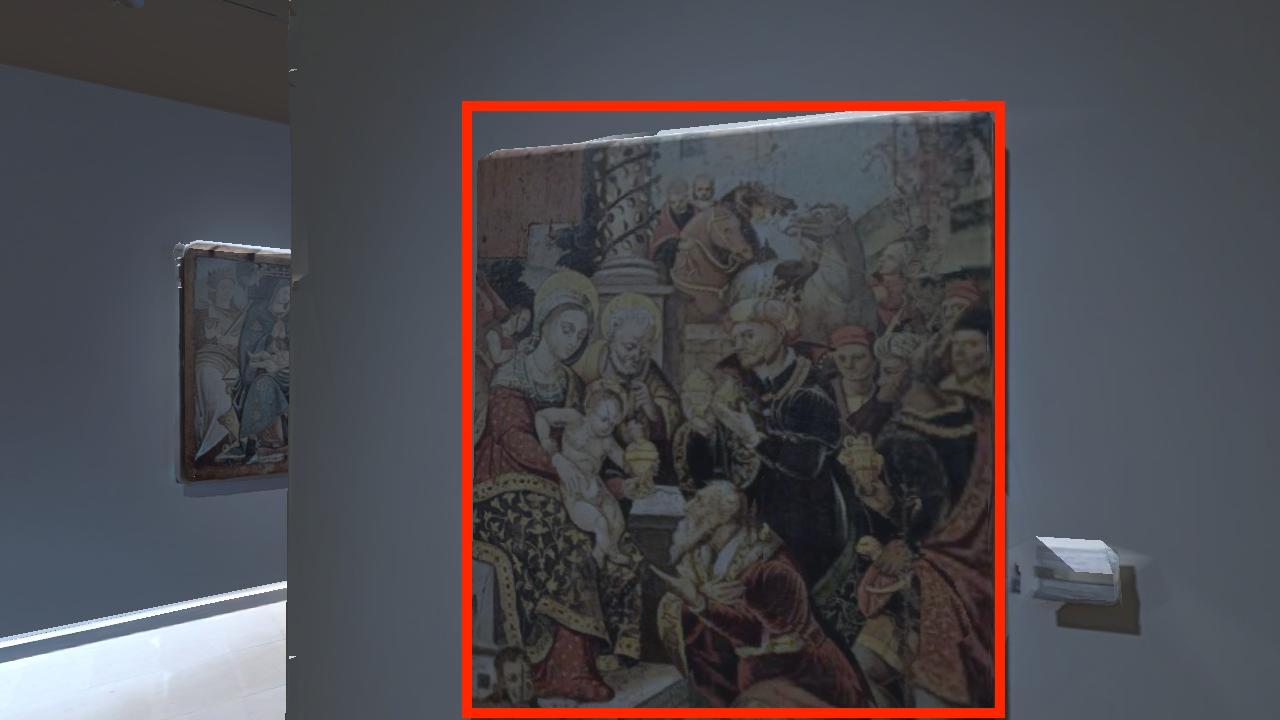}
            \includegraphics[width=.12\textwidth]{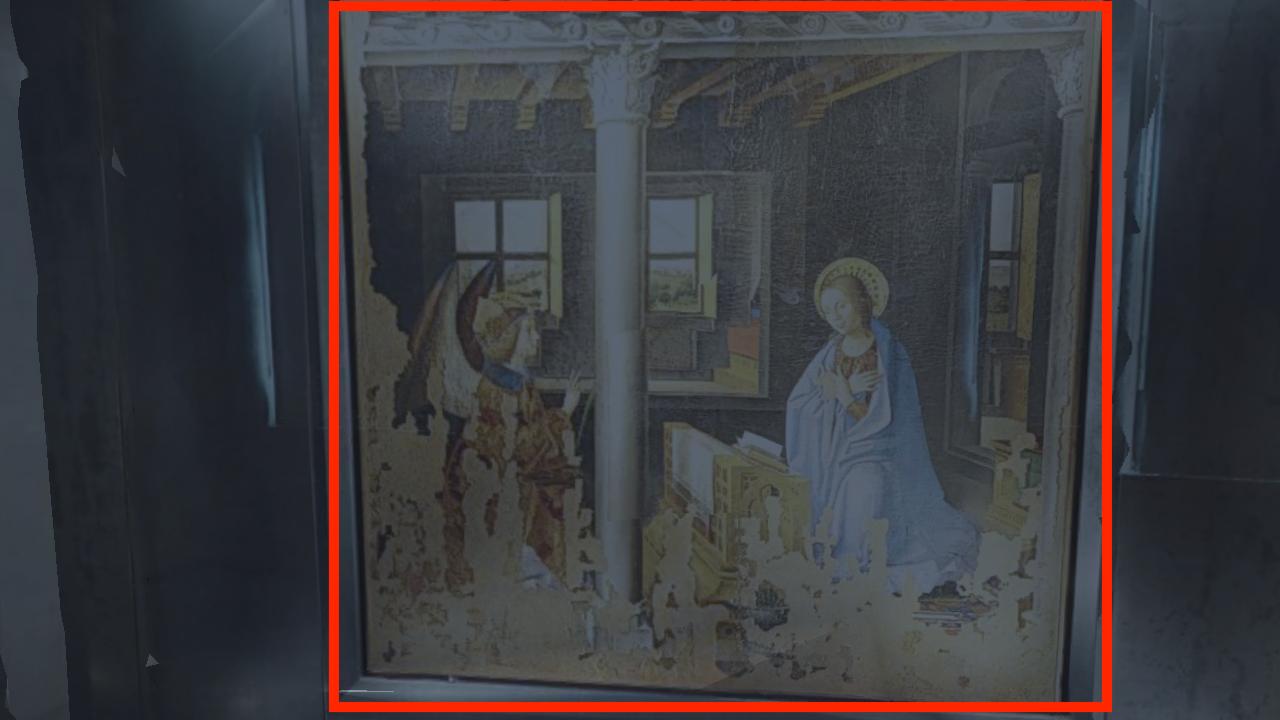}
            \vspace{0.1cm}
            \includegraphics[width=.12\textwidth]{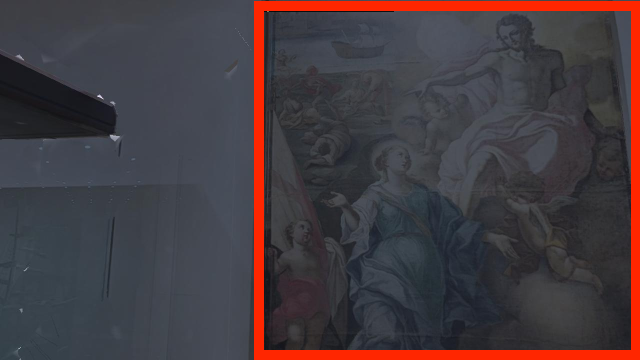}
            \includegraphics[width=.12\textwidth]{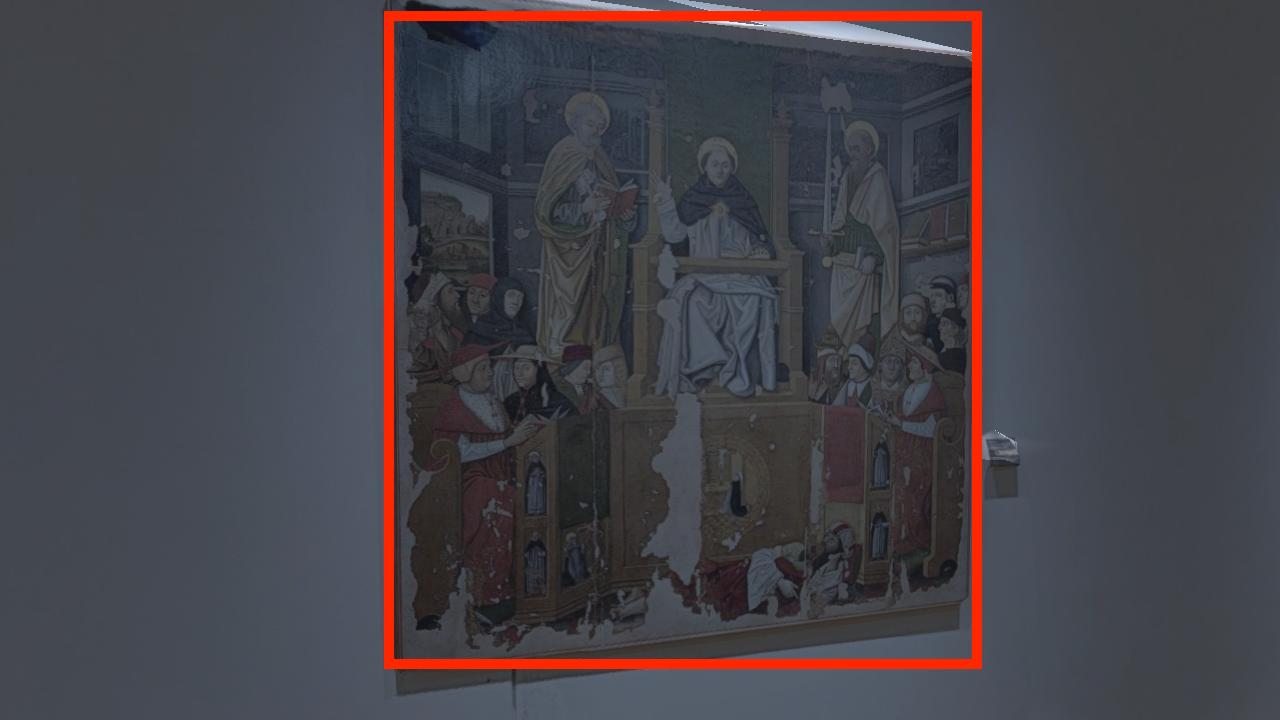}
            \includegraphics[width=.12\textwidth]{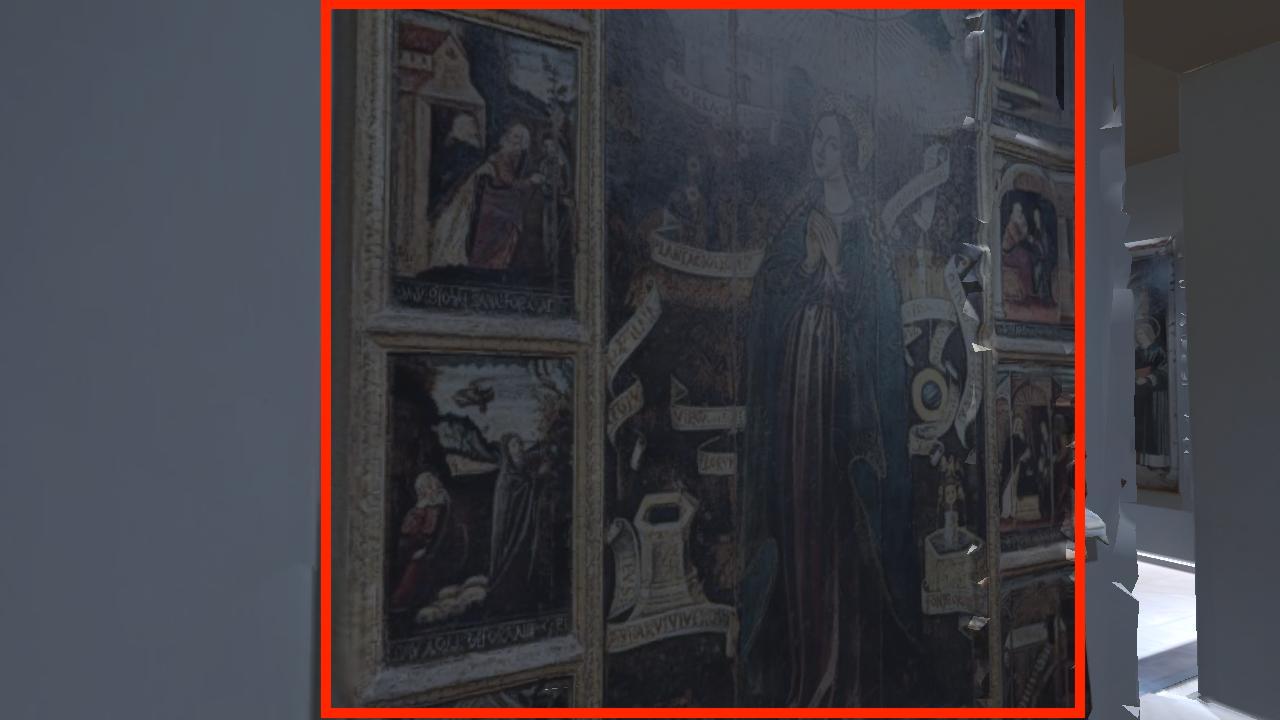}
            \includegraphics[width=.12\textwidth]{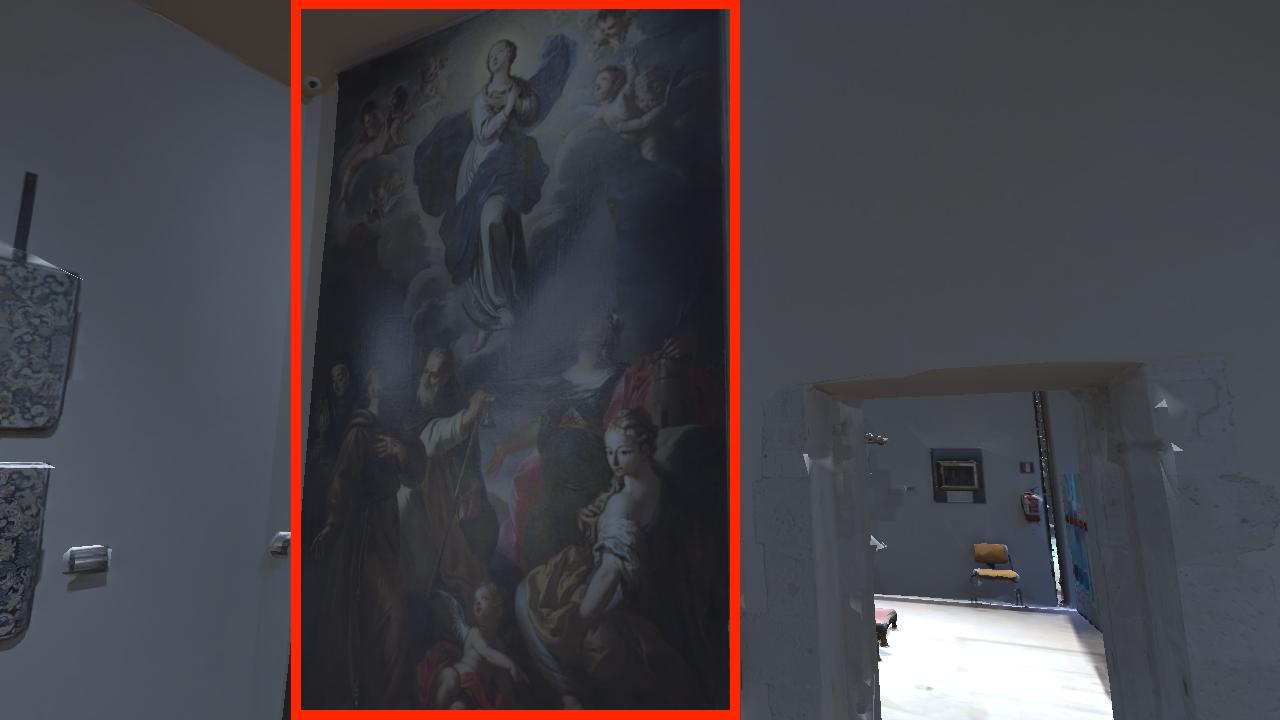}
            \vspace{0.1cm}
            \includegraphics[width=.12\textwidth]{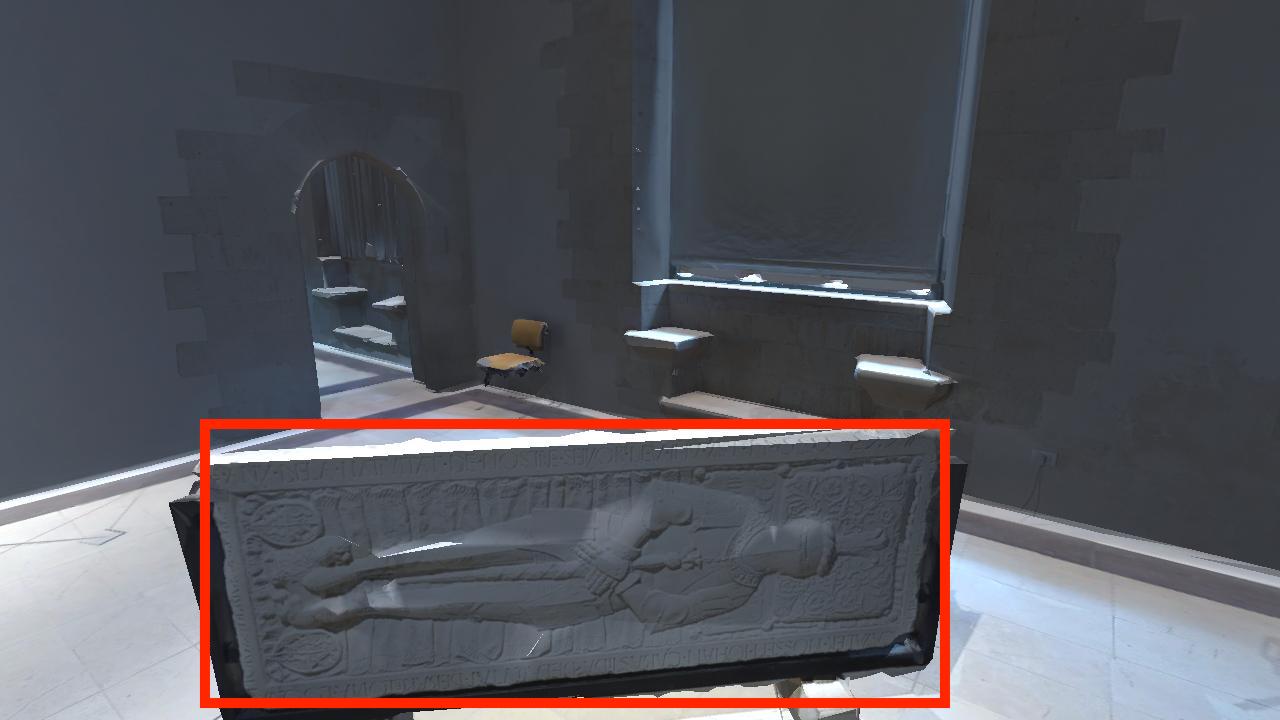}
            \includegraphics[width=.12\textwidth]{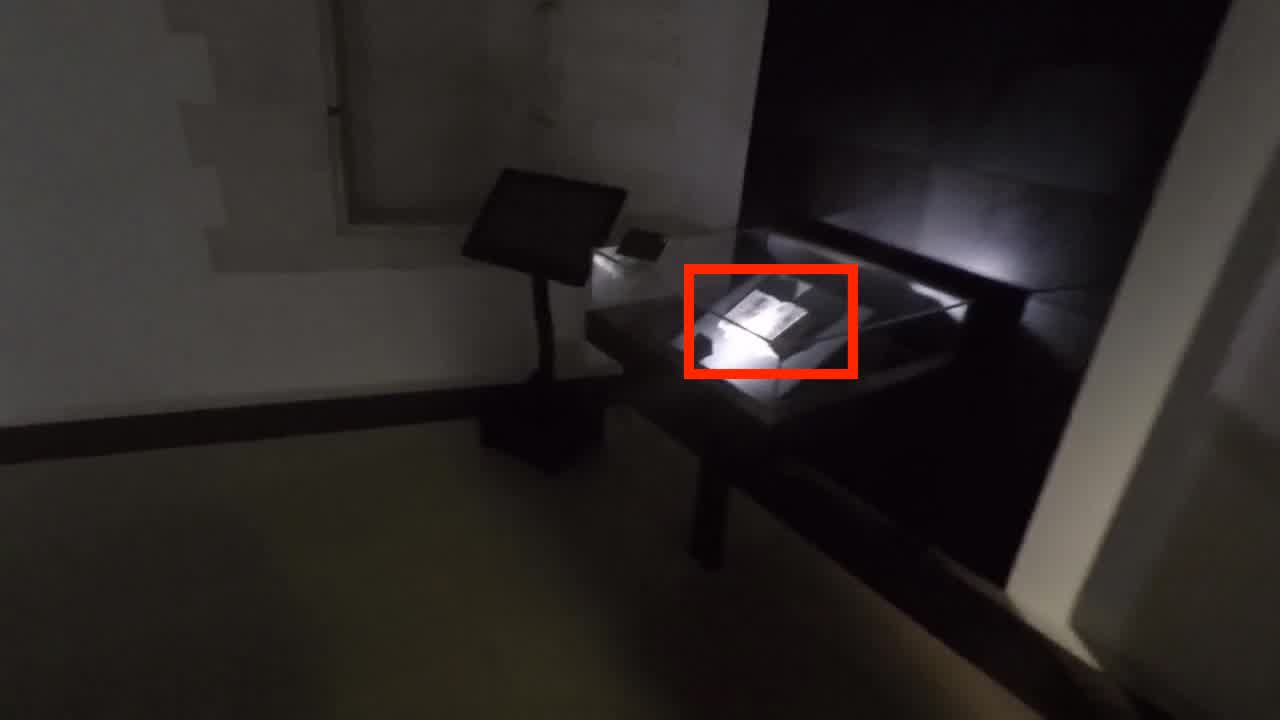}
            \includegraphics[width=.12\textwidth]{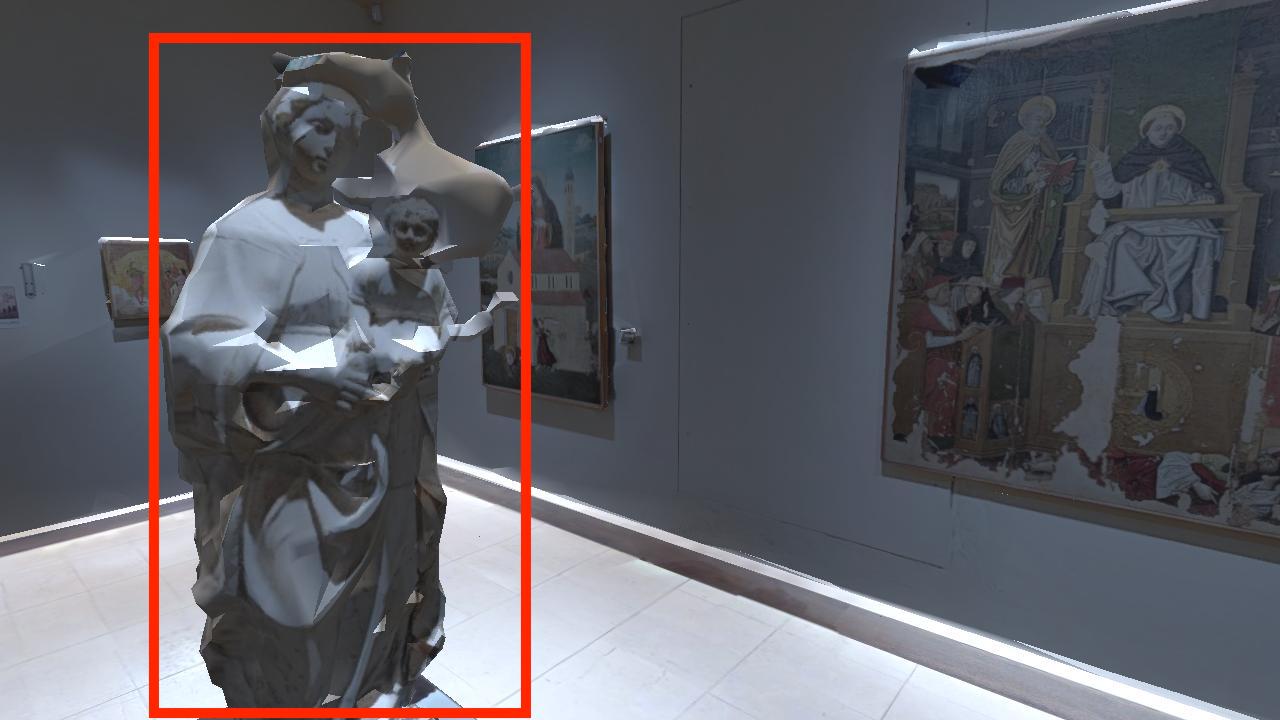}
            \includegraphics[width=.12\textwidth]{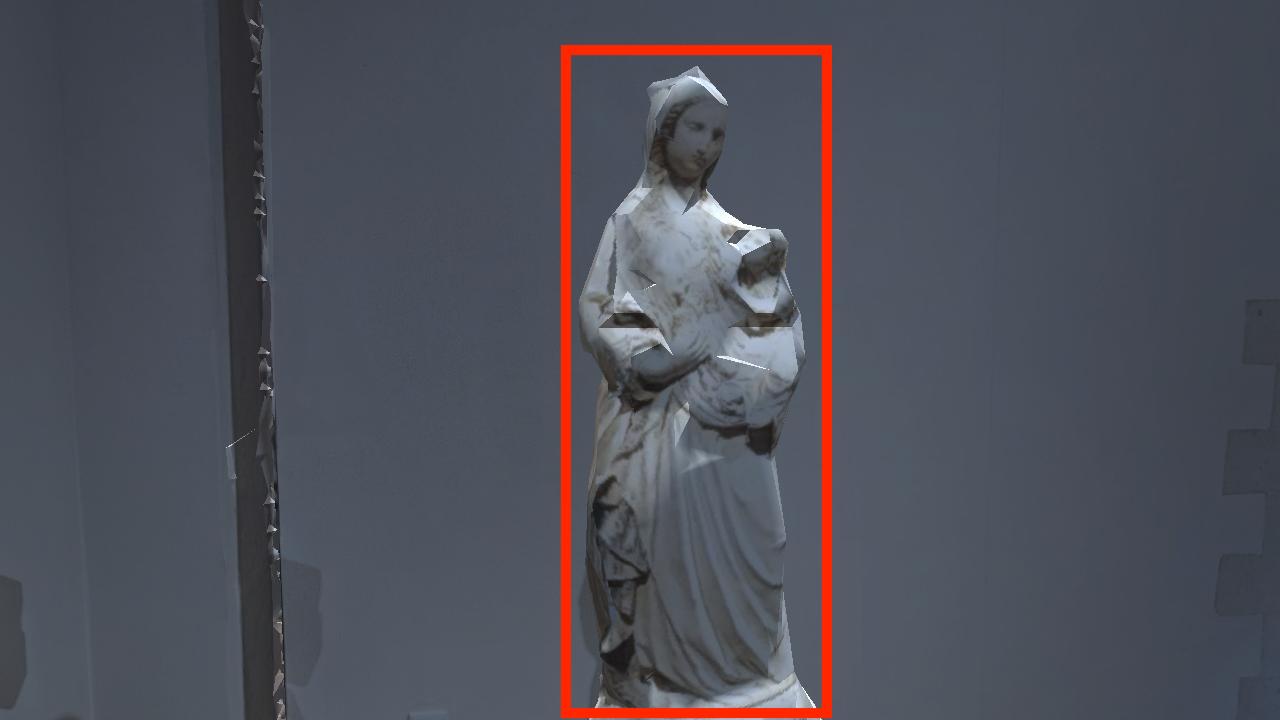}
            \includegraphics[width=.12\textwidth]{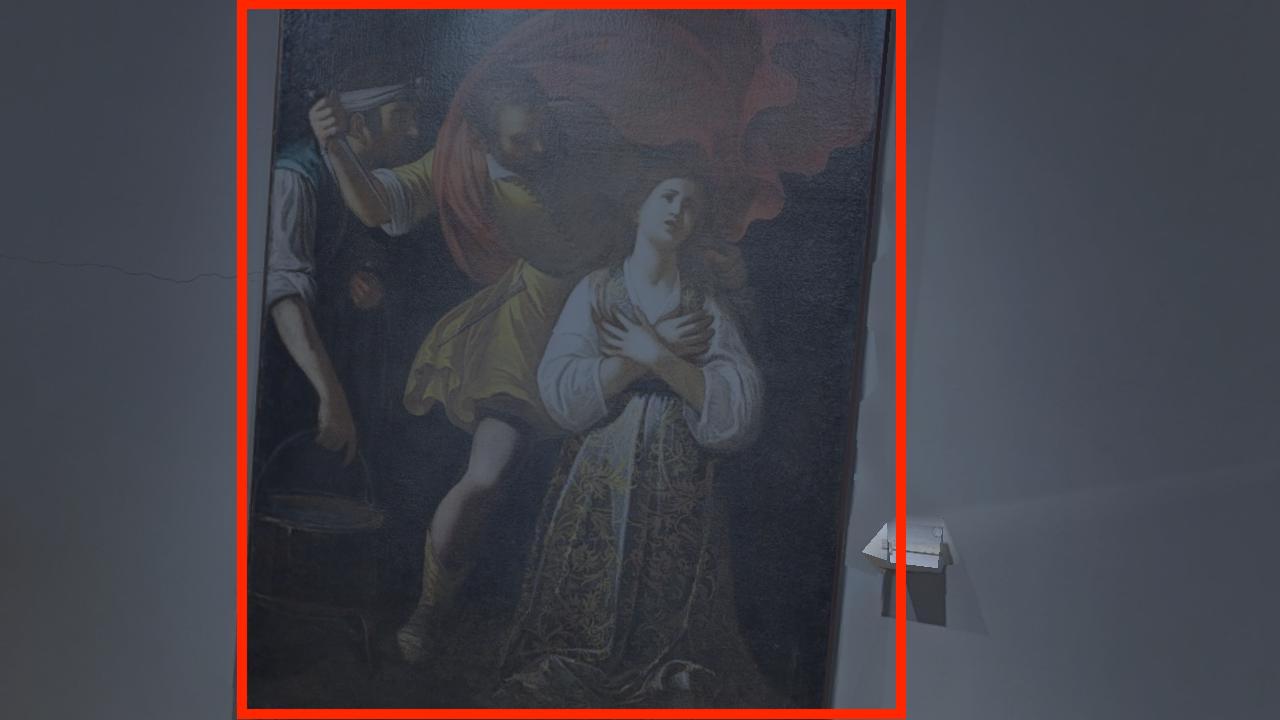}
            \includegraphics[width=.12\textwidth]{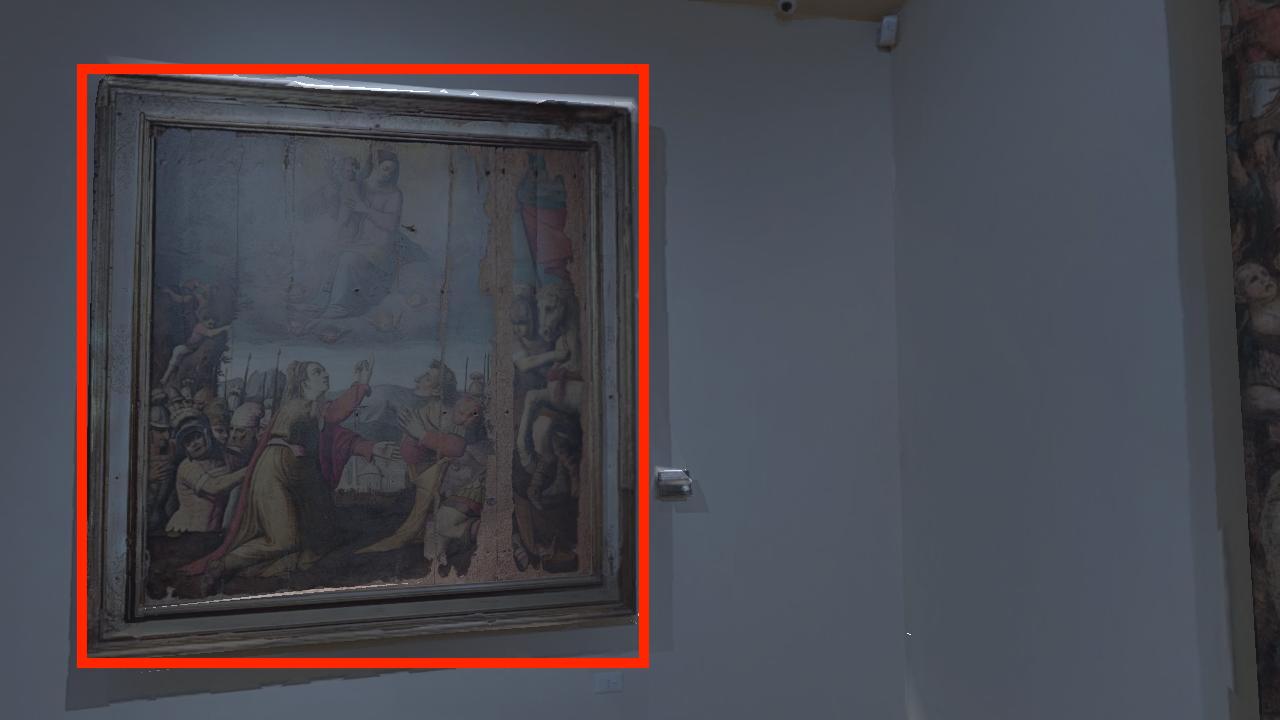}
            \includegraphics[width=.12\textwidth]{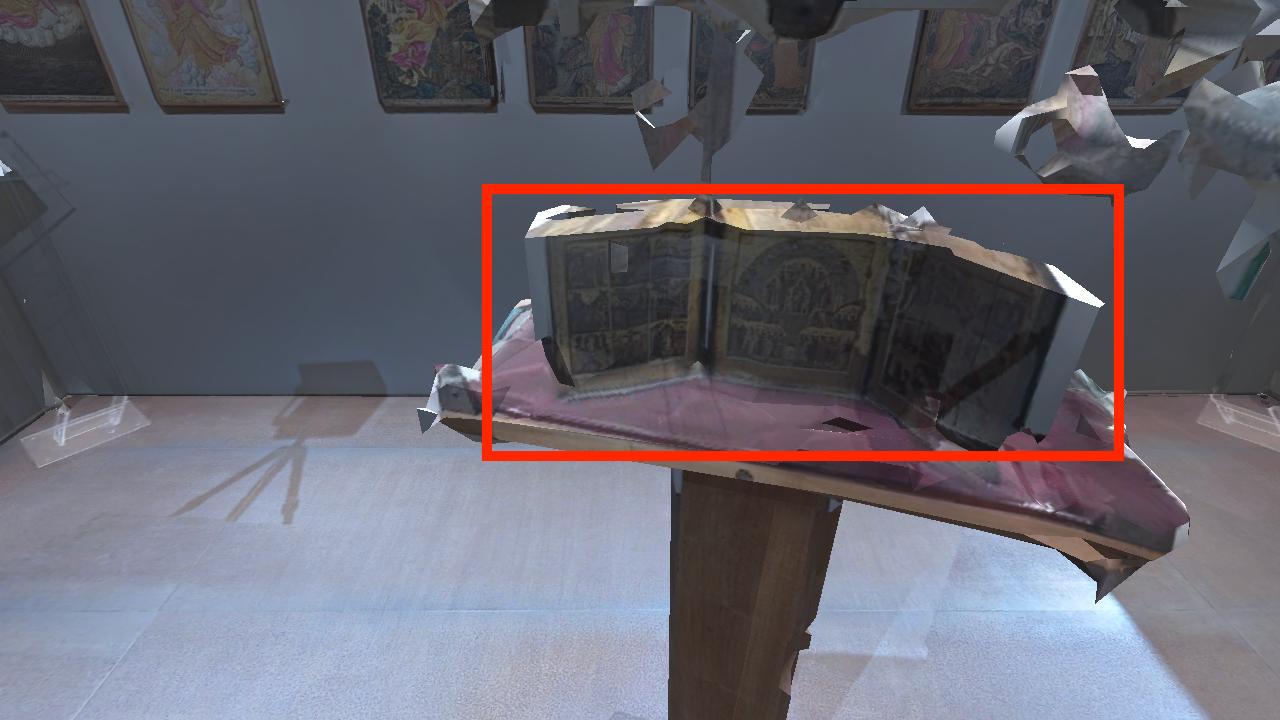}
            \includegraphics[width=.12\textwidth]{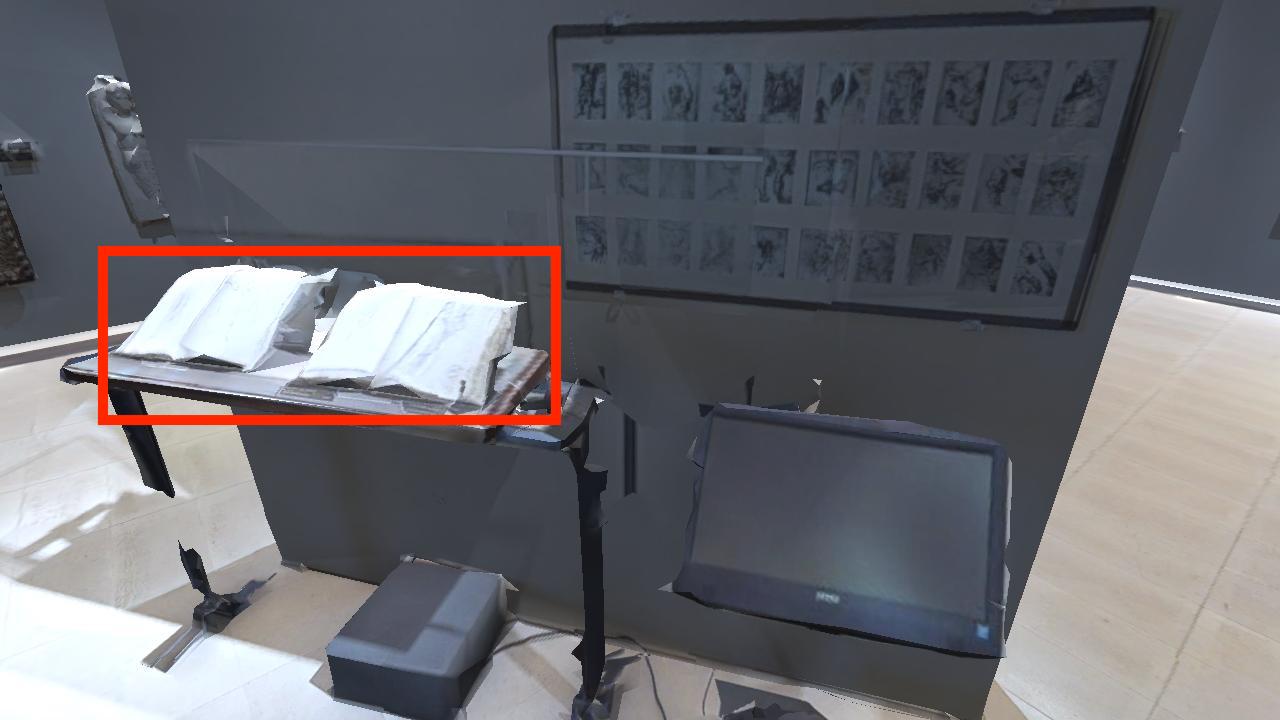}
            \includegraphics[width=.12\textwidth]{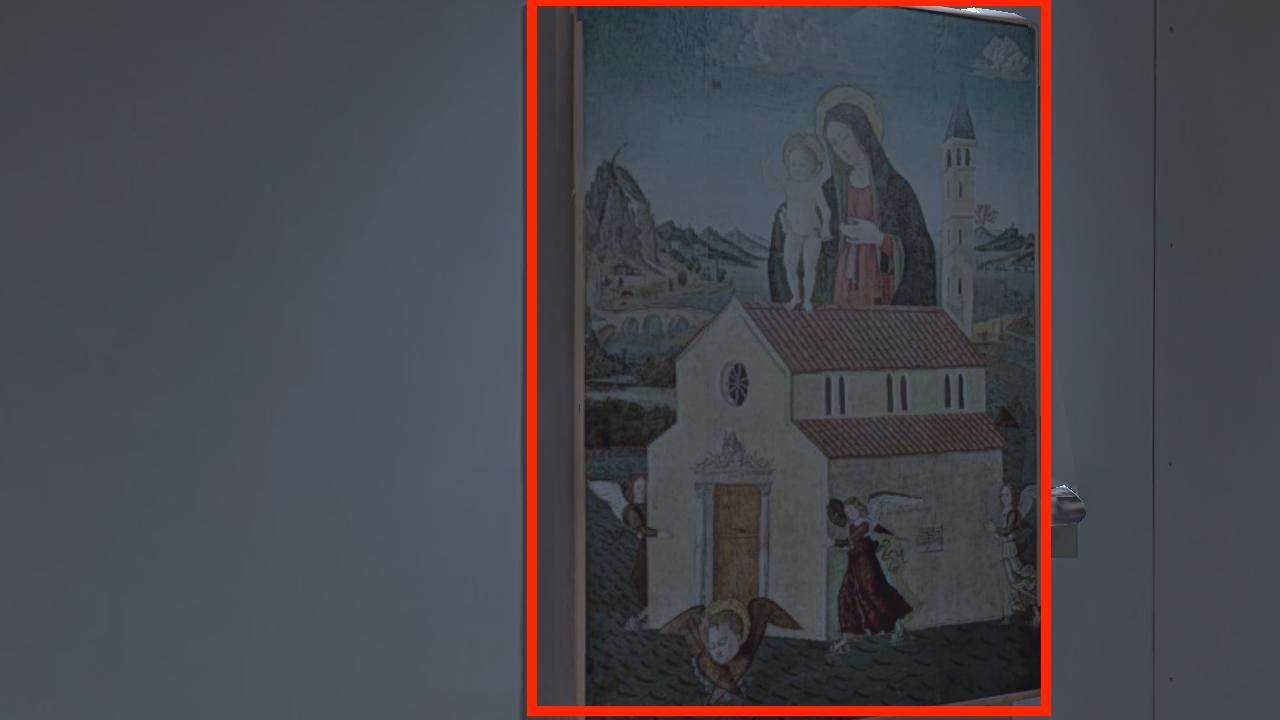}
            \includegraphics[width=.12\textwidth]{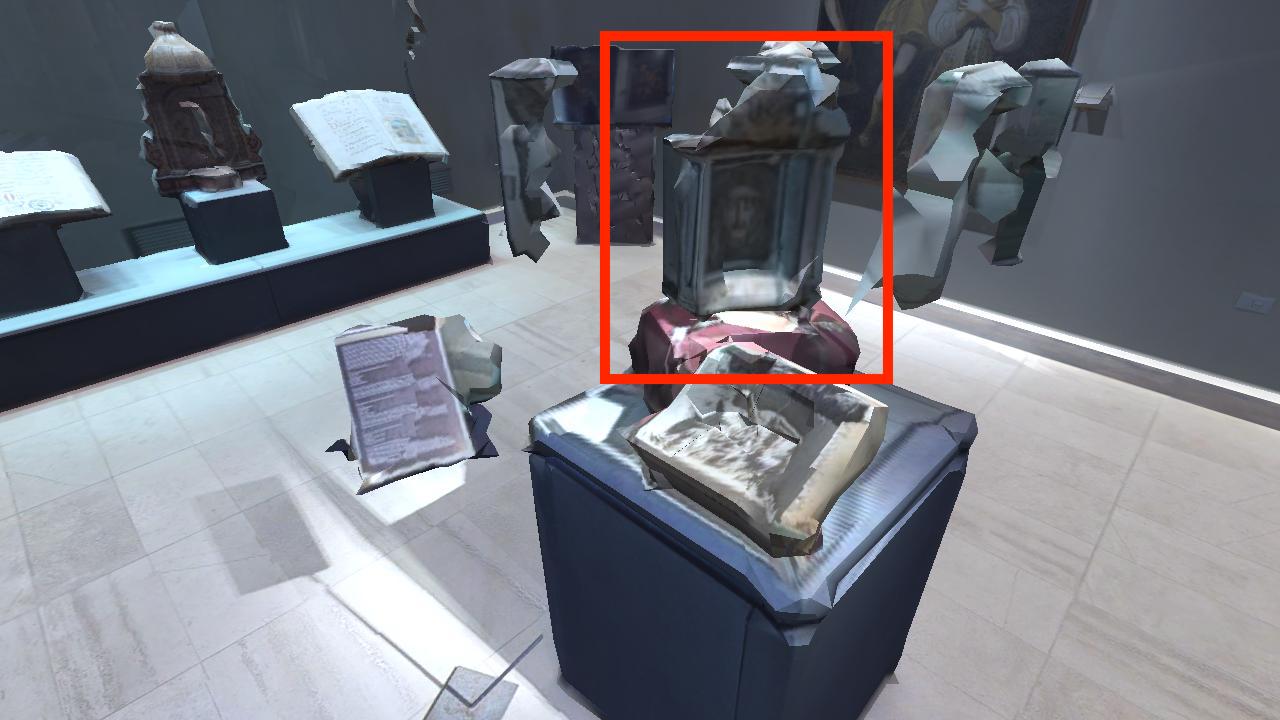}
            \subcaption{Synthetic images.}
            \includegraphics[width=.12\textwidth]{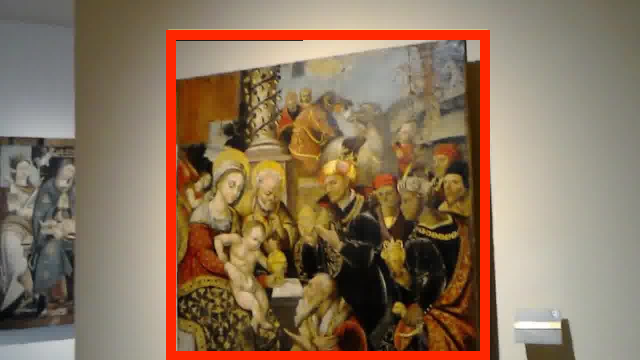}
            \includegraphics[width=.12\textwidth]{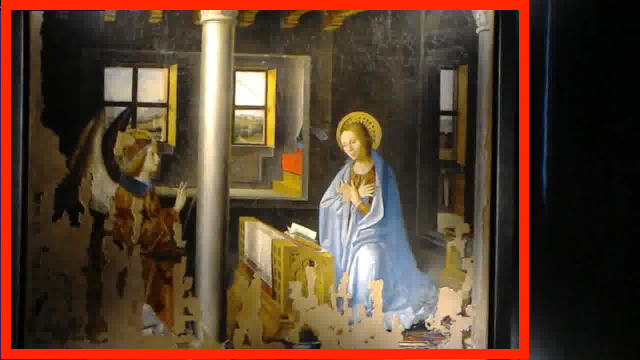}
            \vspace{0.1cm}
            \includegraphics[width=.12\textwidth]{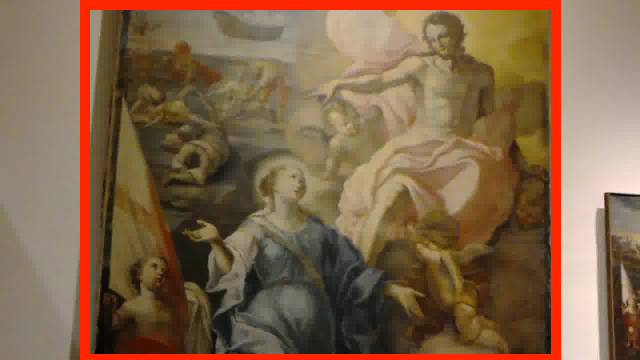}
            \includegraphics[width=.12\textwidth]{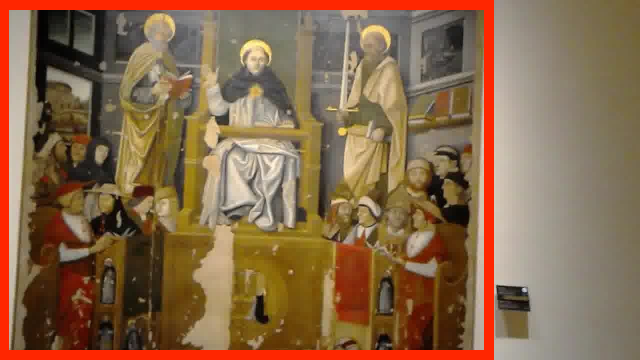}
            \includegraphics[width=.12\textwidth]{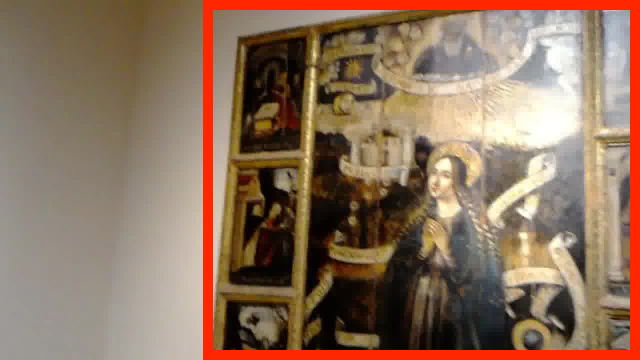}
            \includegraphics[width=.12\textwidth]{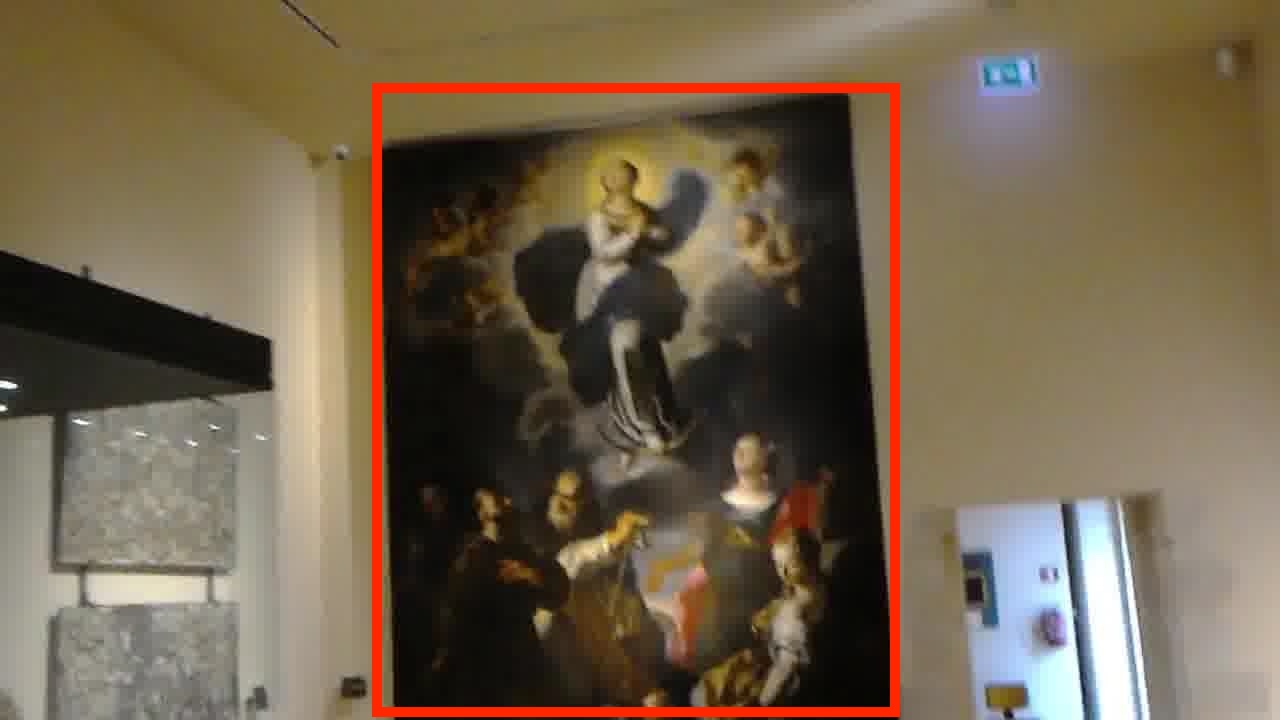}
            \includegraphics[width=.12\textwidth]{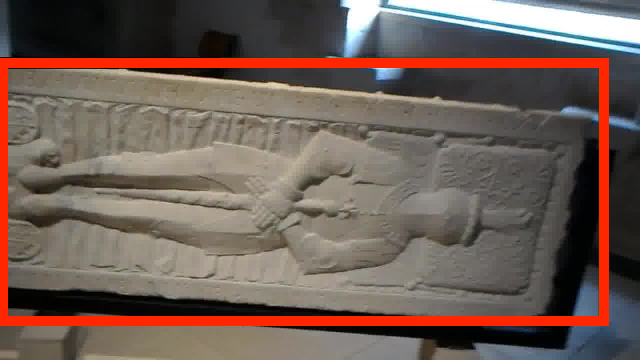}
            \vspace{0.1cm}
            \includegraphics[width=.12\textwidth]{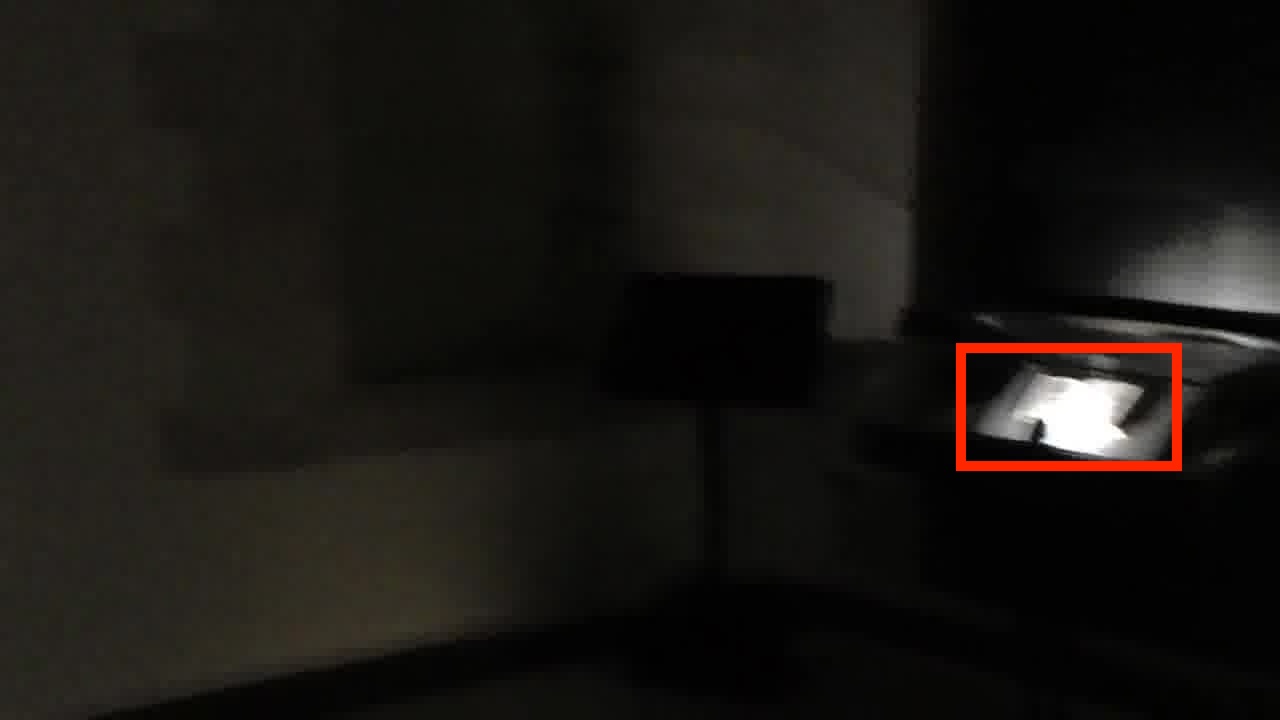}
            \includegraphics[width=.12\textwidth]{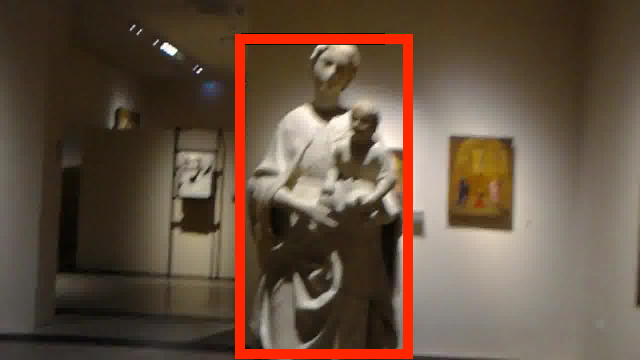}
            \includegraphics[width=.12\textwidth]{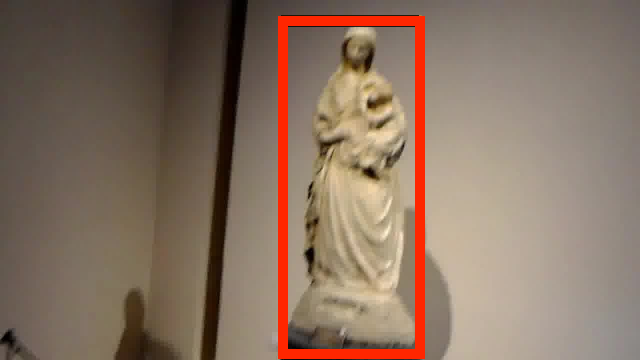}
            \includegraphics[width=.12\textwidth]{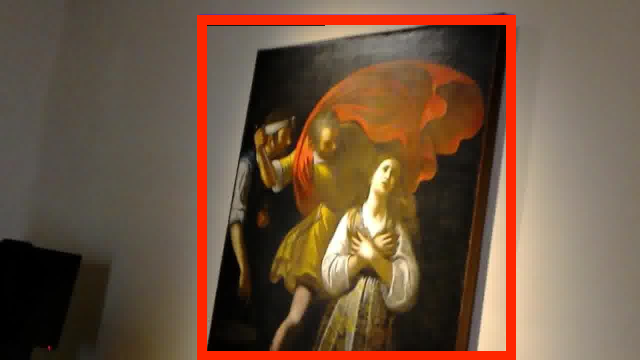}
            \includegraphics[width=.12\textwidth]{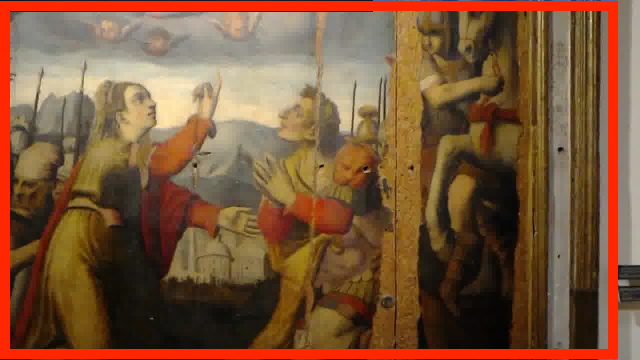}
            \includegraphics[width=.12\textwidth]{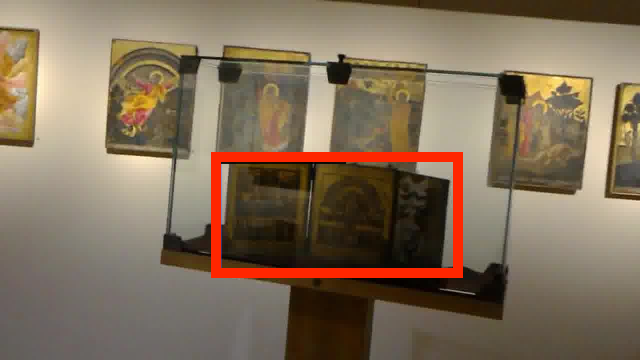}
            \includegraphics[width=.12\textwidth]{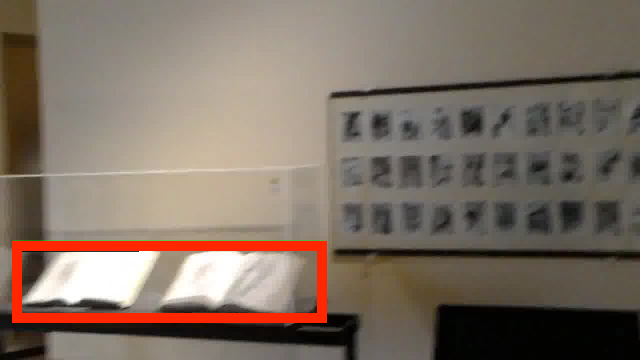}
            \includegraphics[width=.12\textwidth]{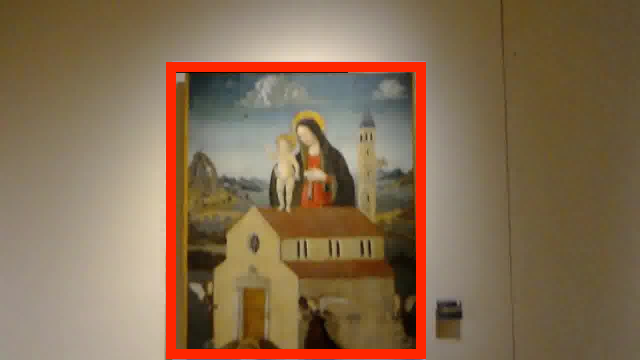}
            \includegraphics[width=.12\textwidth]{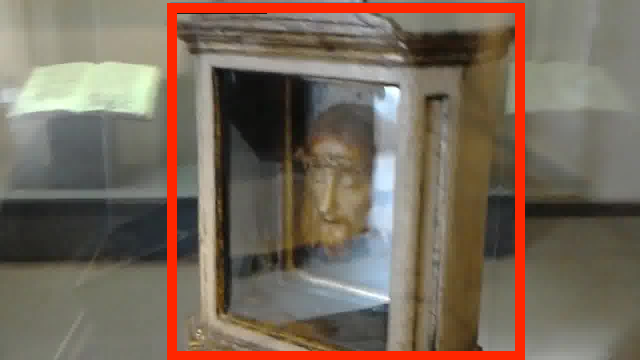}
            \subcaption{Images acquired with HoloLens.}
            \includegraphics[width=.12\textwidth]{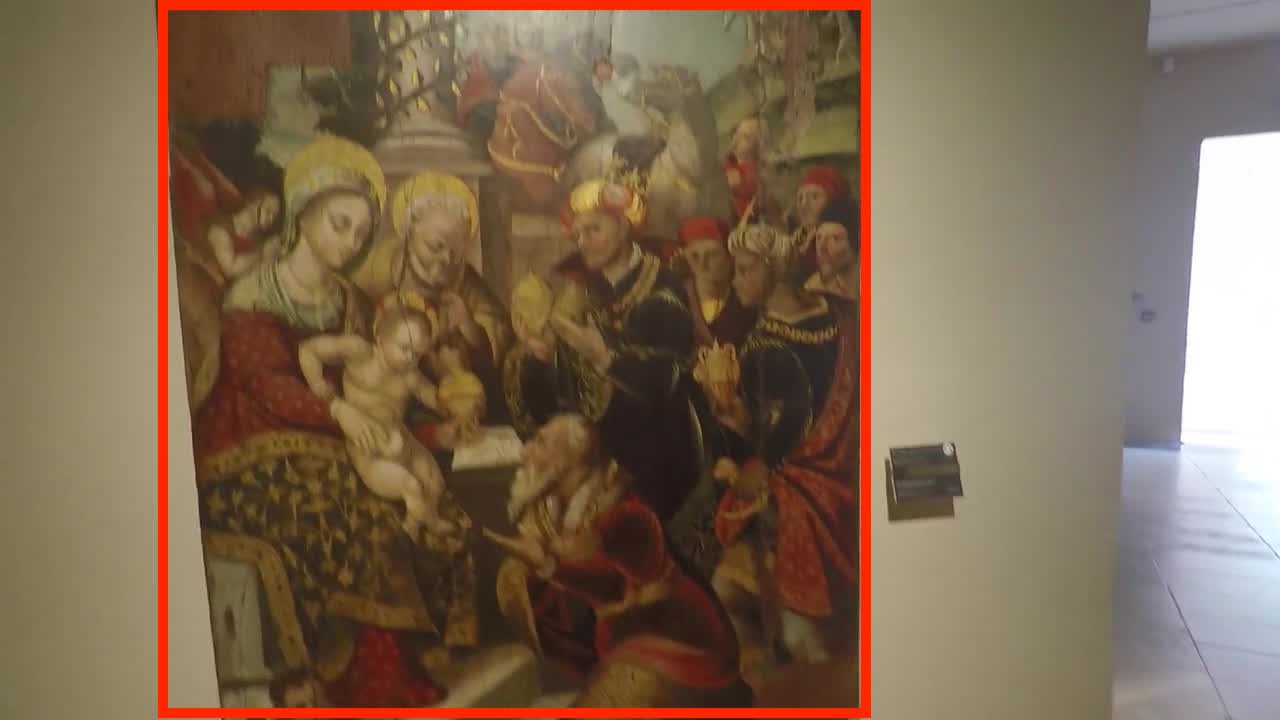}
            \includegraphics[width=.12\textwidth]{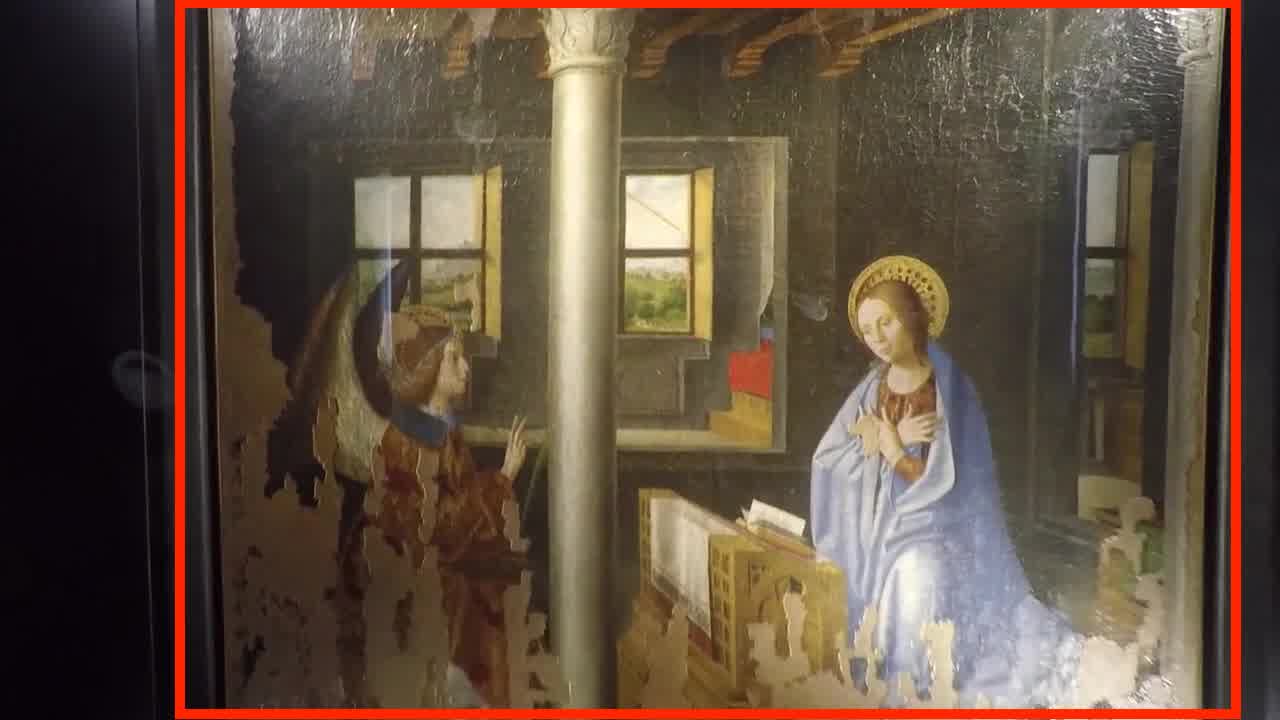}
            \vspace{0.1cm}
            \includegraphics[width=.12\textwidth]{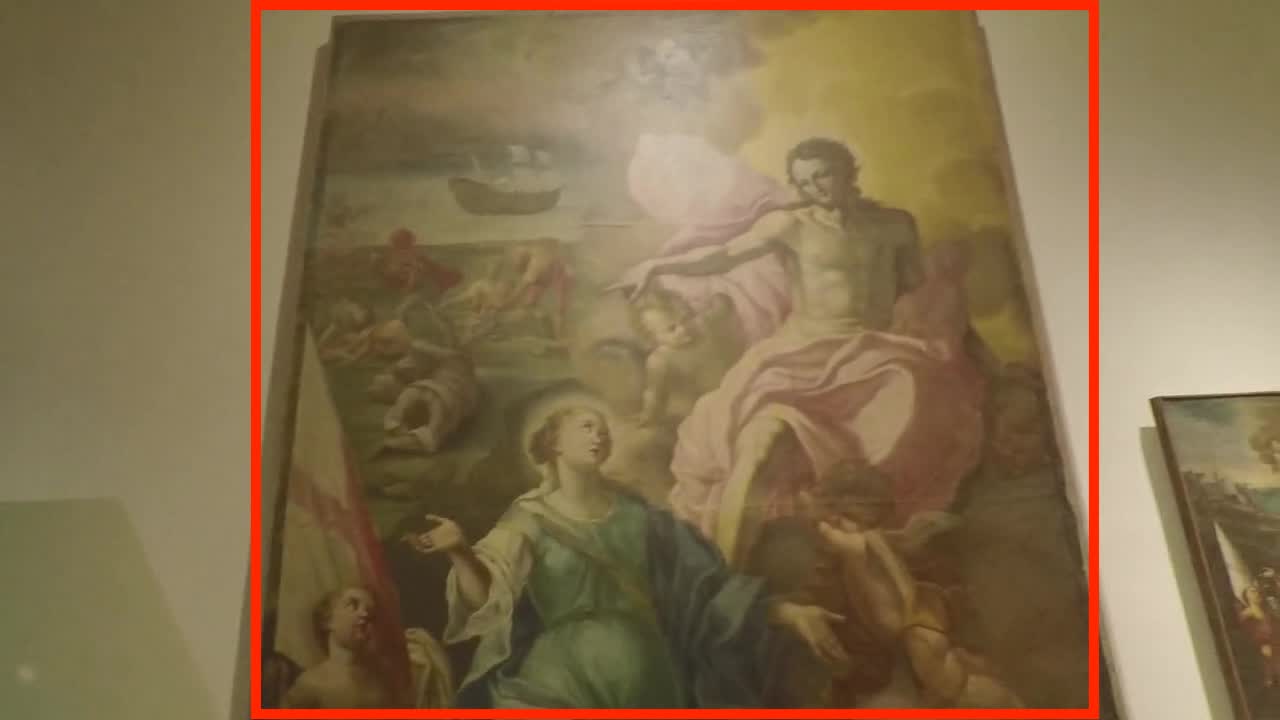}
            \includegraphics[width=.12\textwidth]{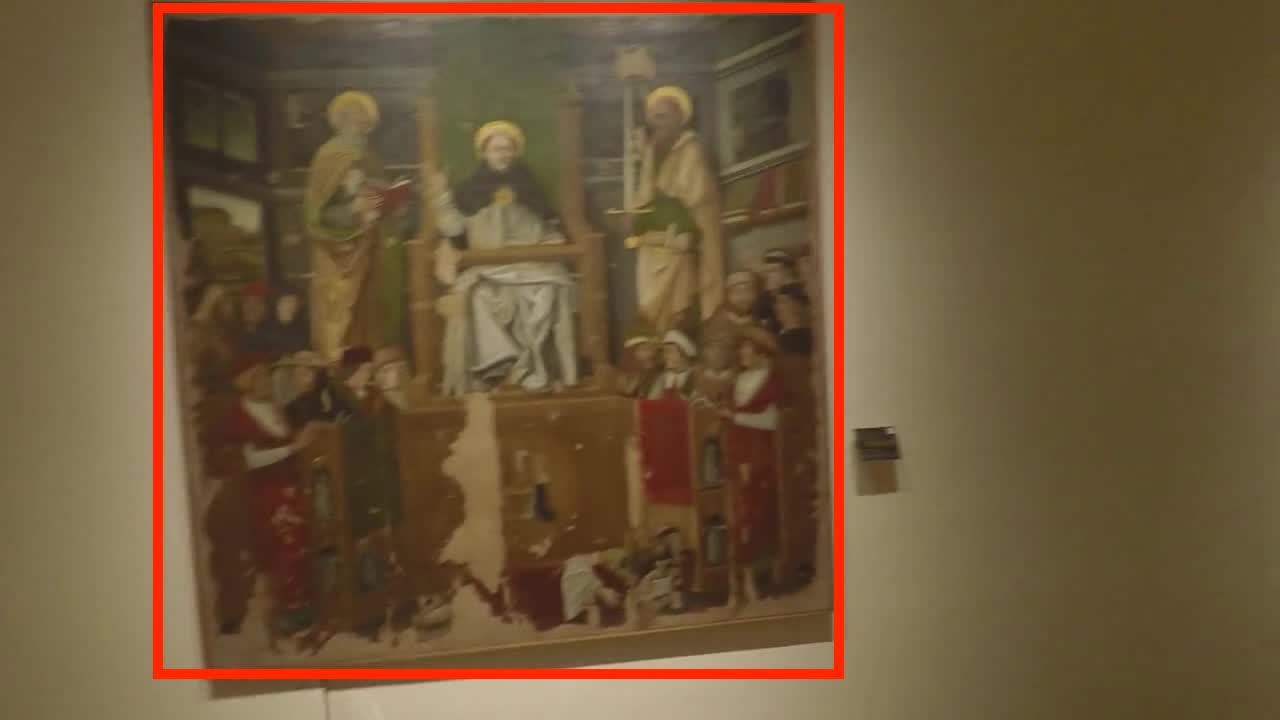}
            \includegraphics[width=.12\textwidth]{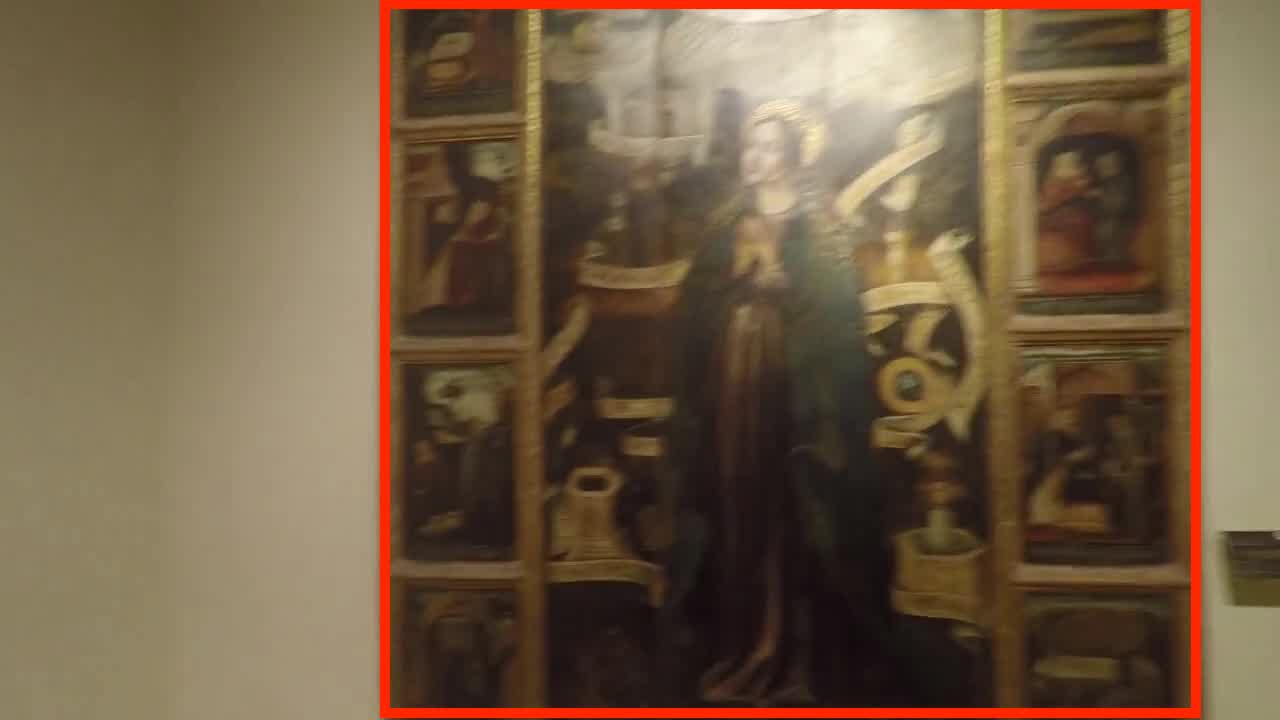}
            \includegraphics[width=.12\textwidth]{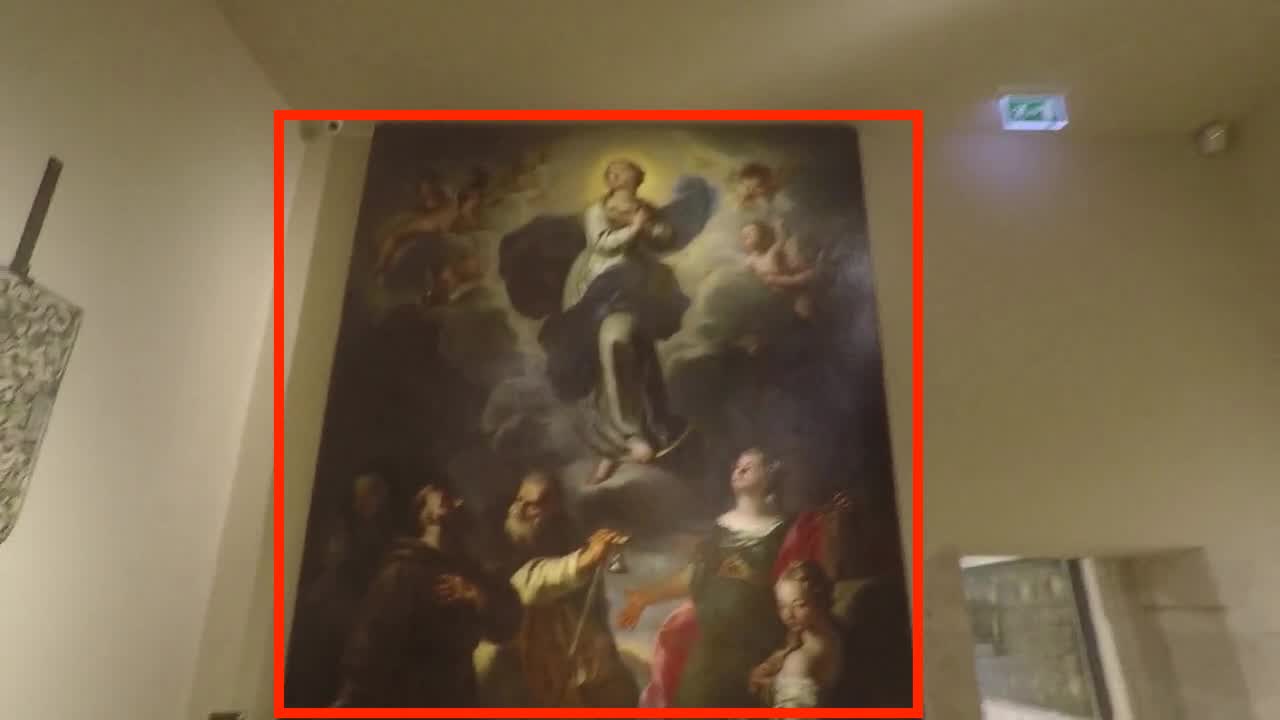}
            \includegraphics[width=.12\textwidth]{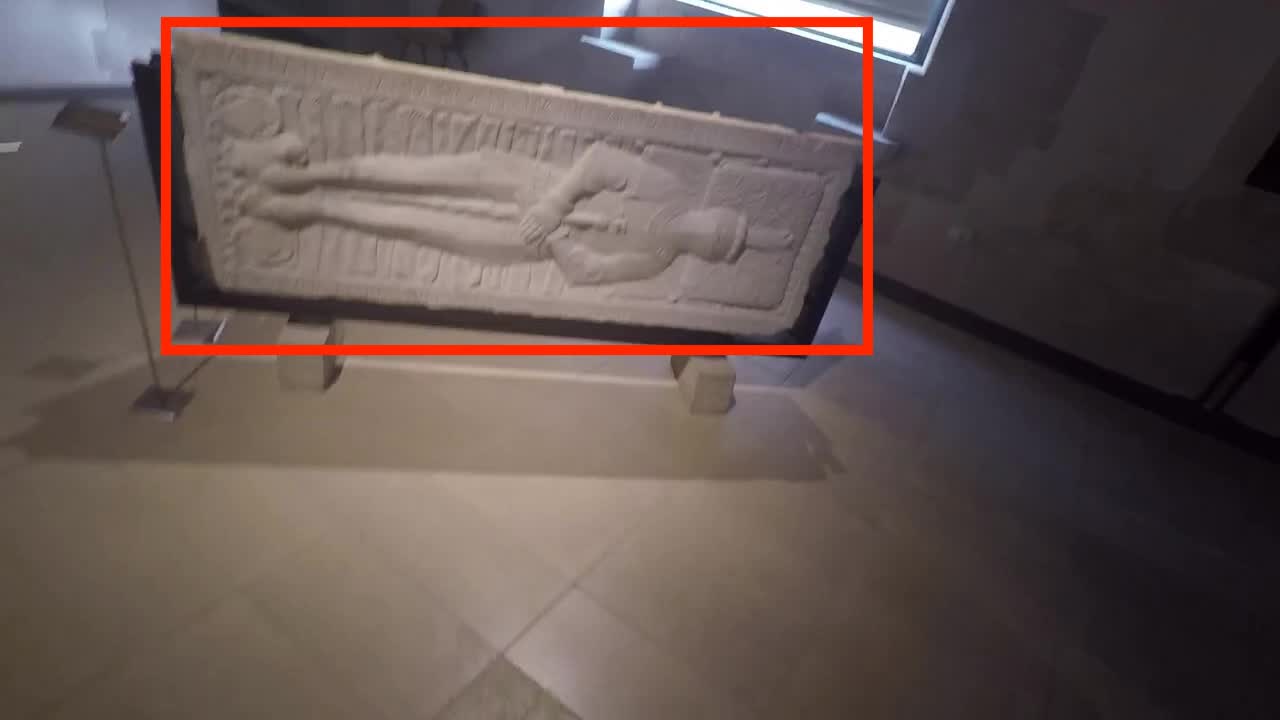}
            \vspace{0.1cm}
            \includegraphics[width=.12\textwidth]{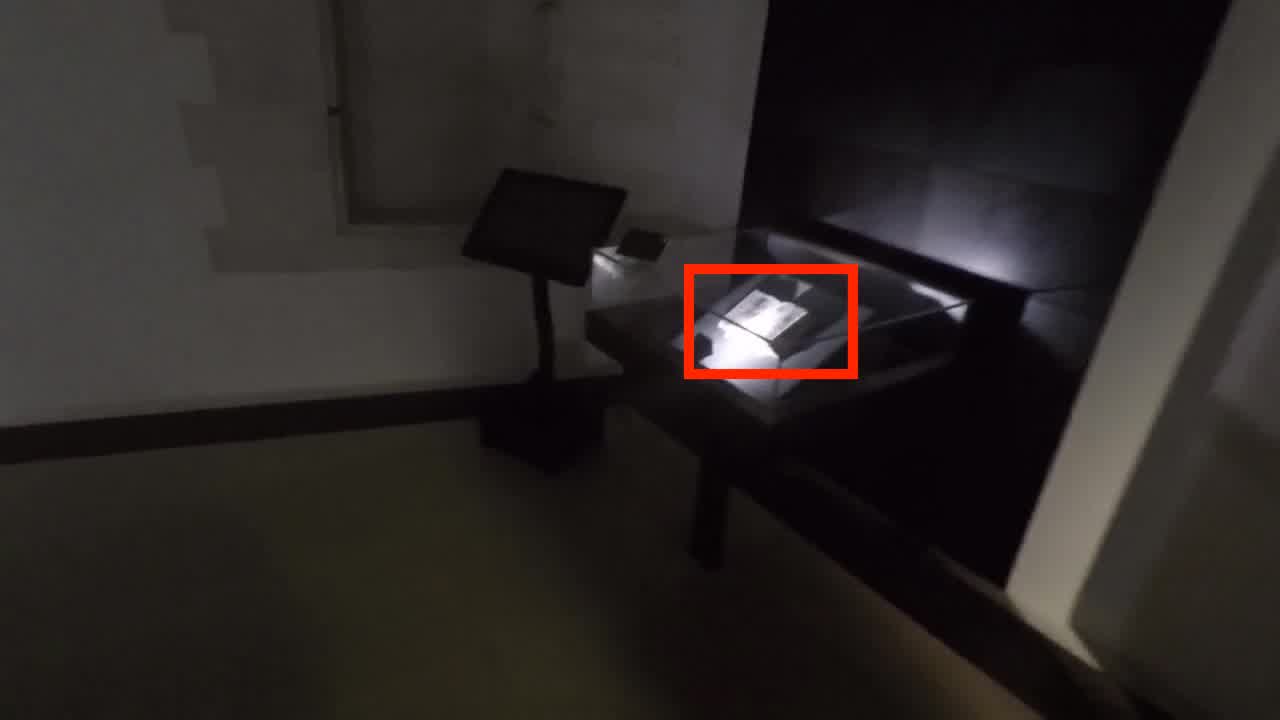}
            \includegraphics[width=.12\textwidth]{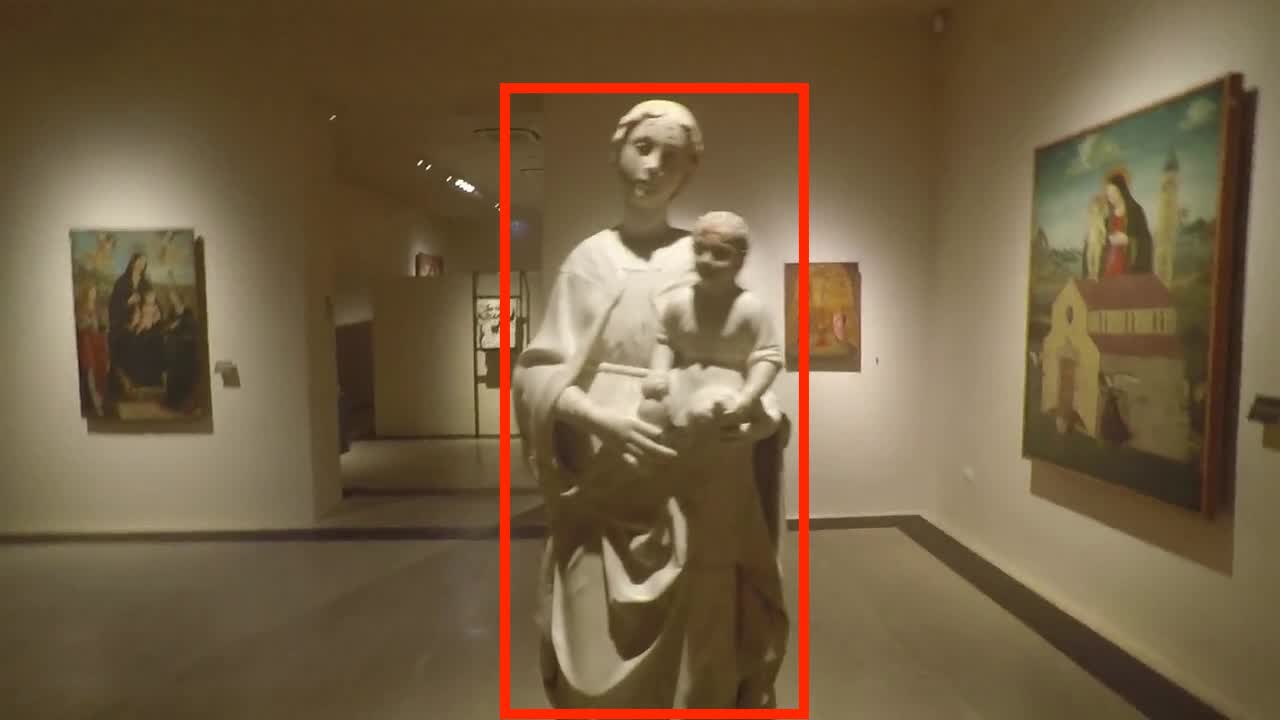}
            \includegraphics[width=.12\textwidth]{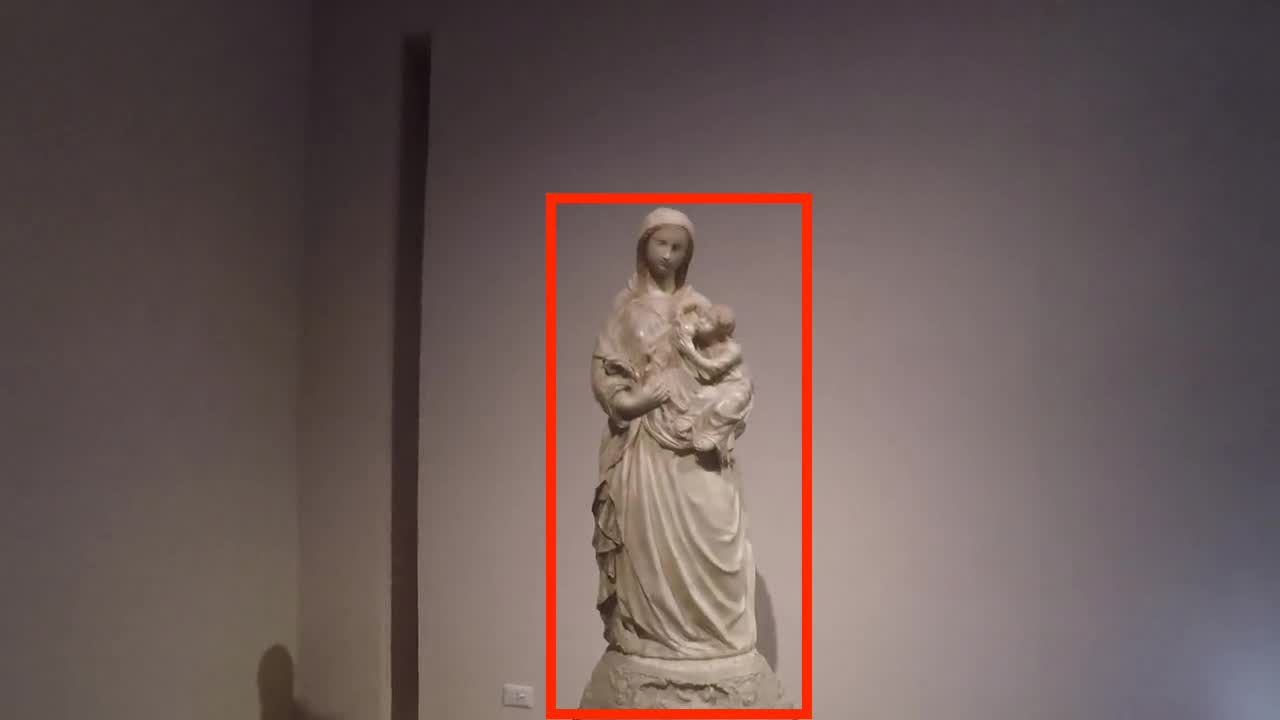}
            \includegraphics[width=.12\textwidth]{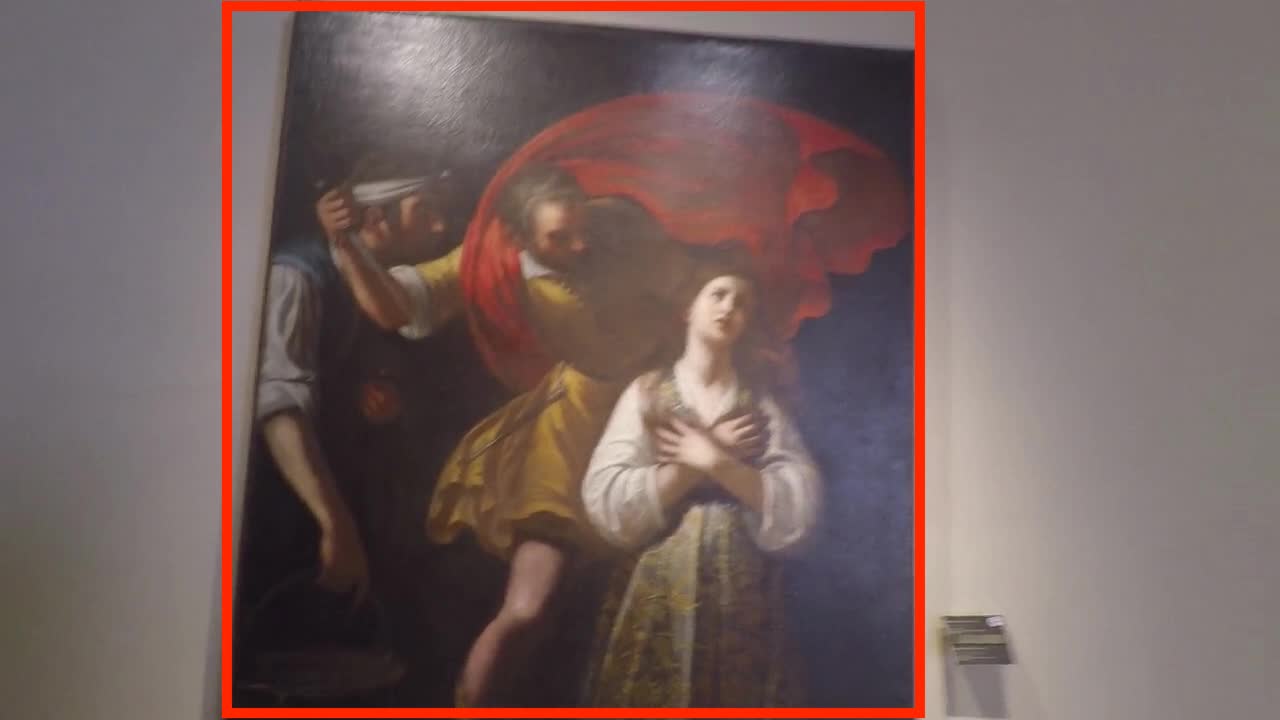}
            \includegraphics[width=.12\textwidth]{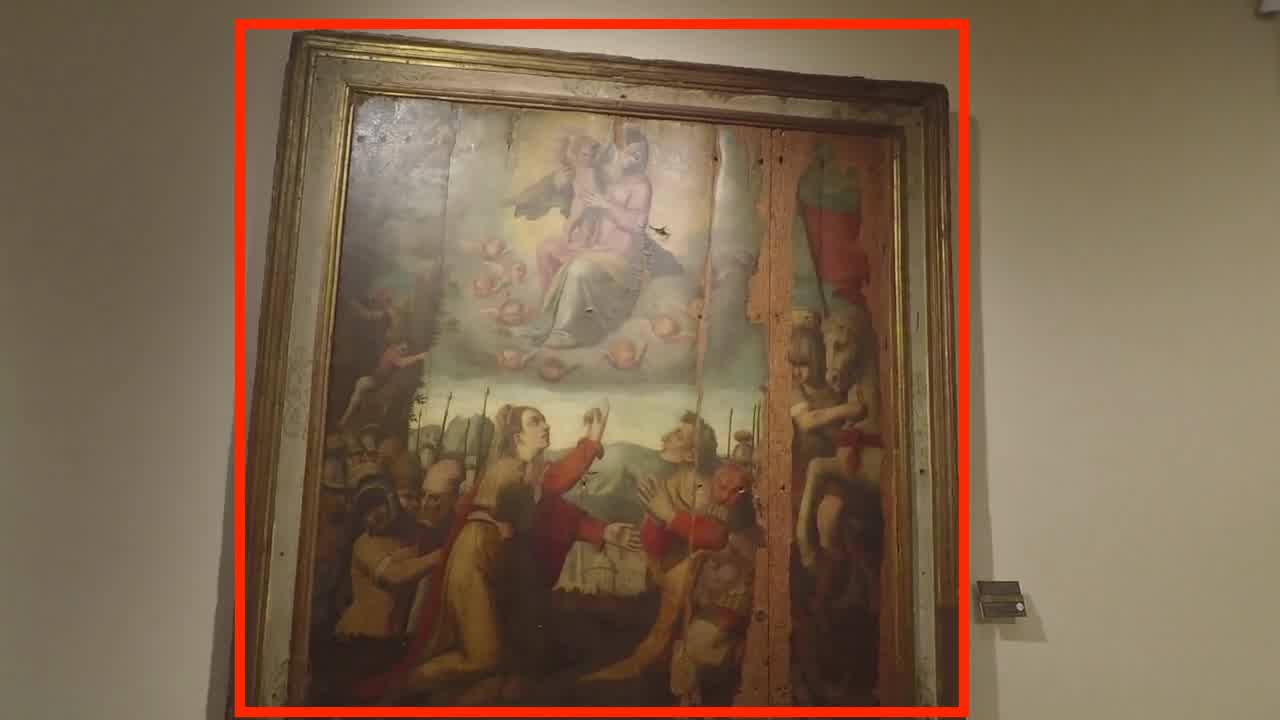}
            \includegraphics[width=.12\textwidth]{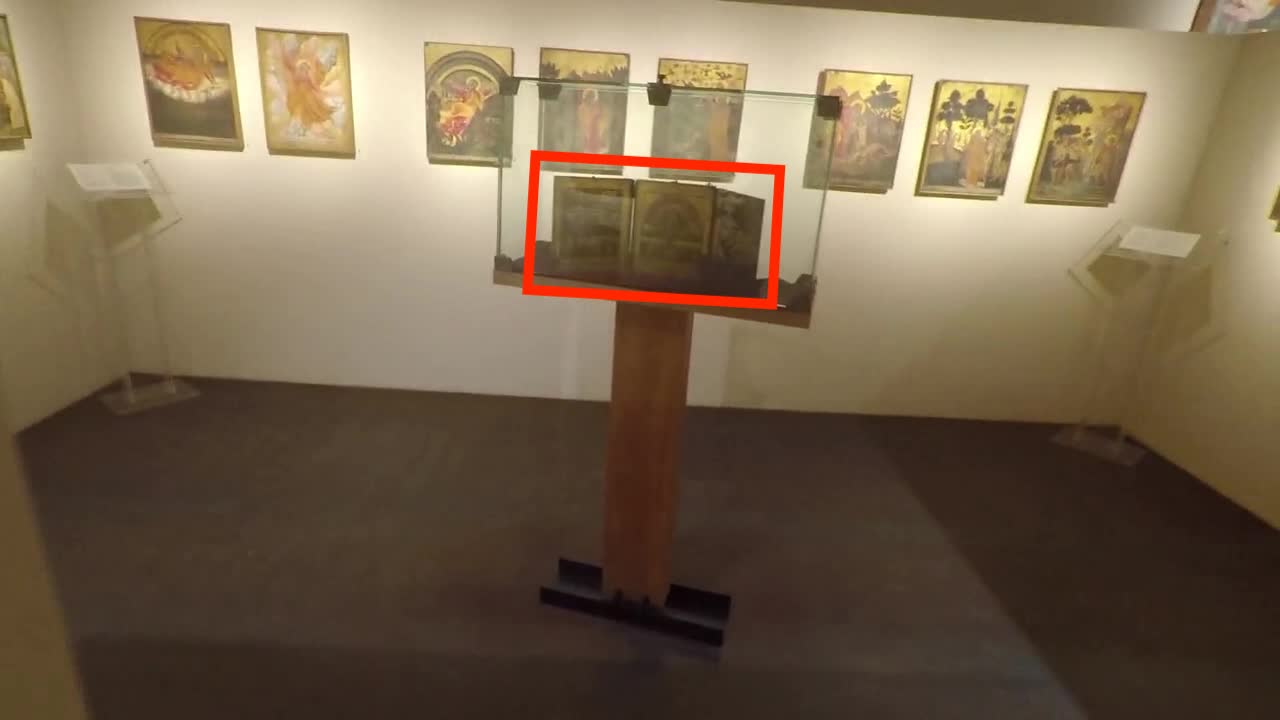}
            \includegraphics[width=.12\textwidth]{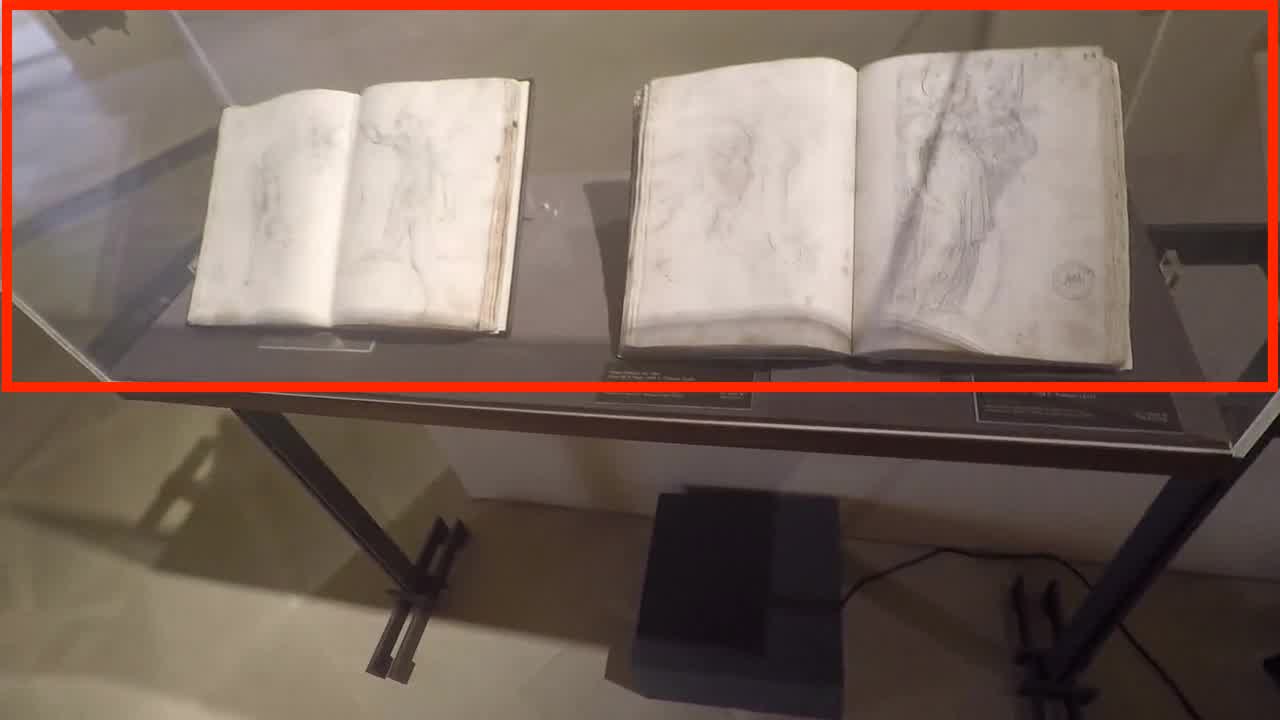}
            \includegraphics[width=.12\textwidth]{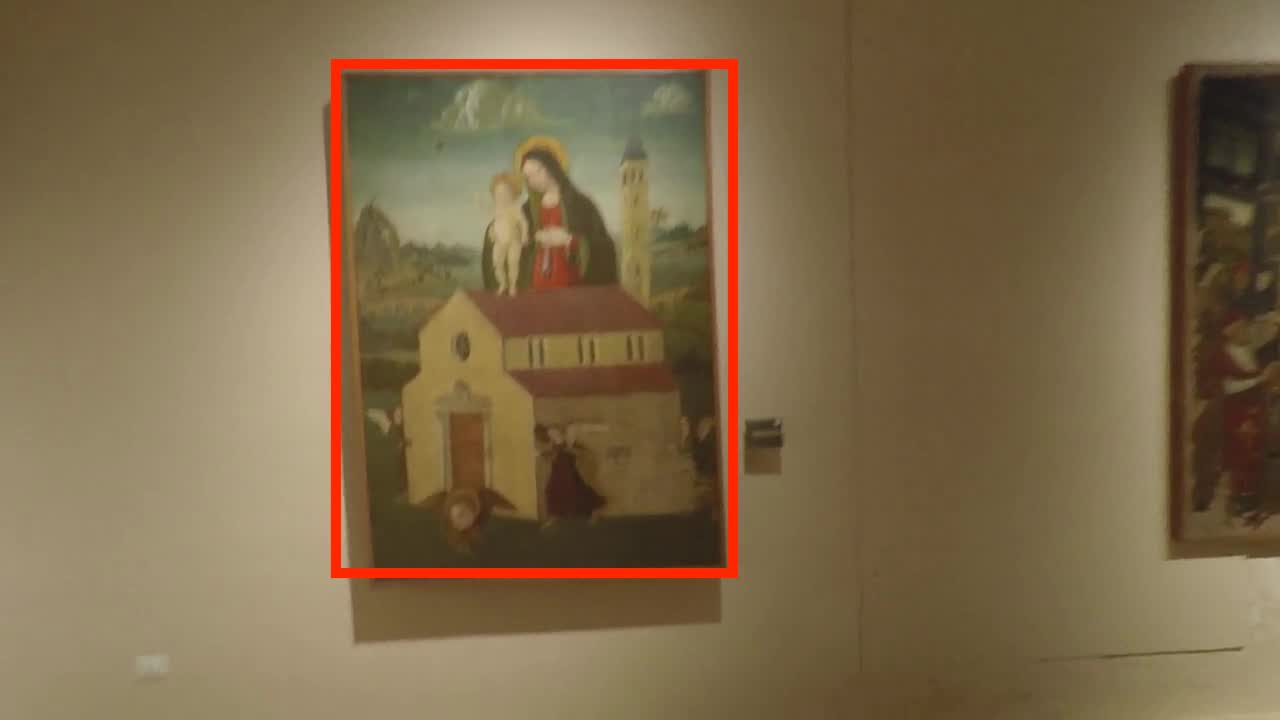}
            \includegraphics[width=.12\textwidth]{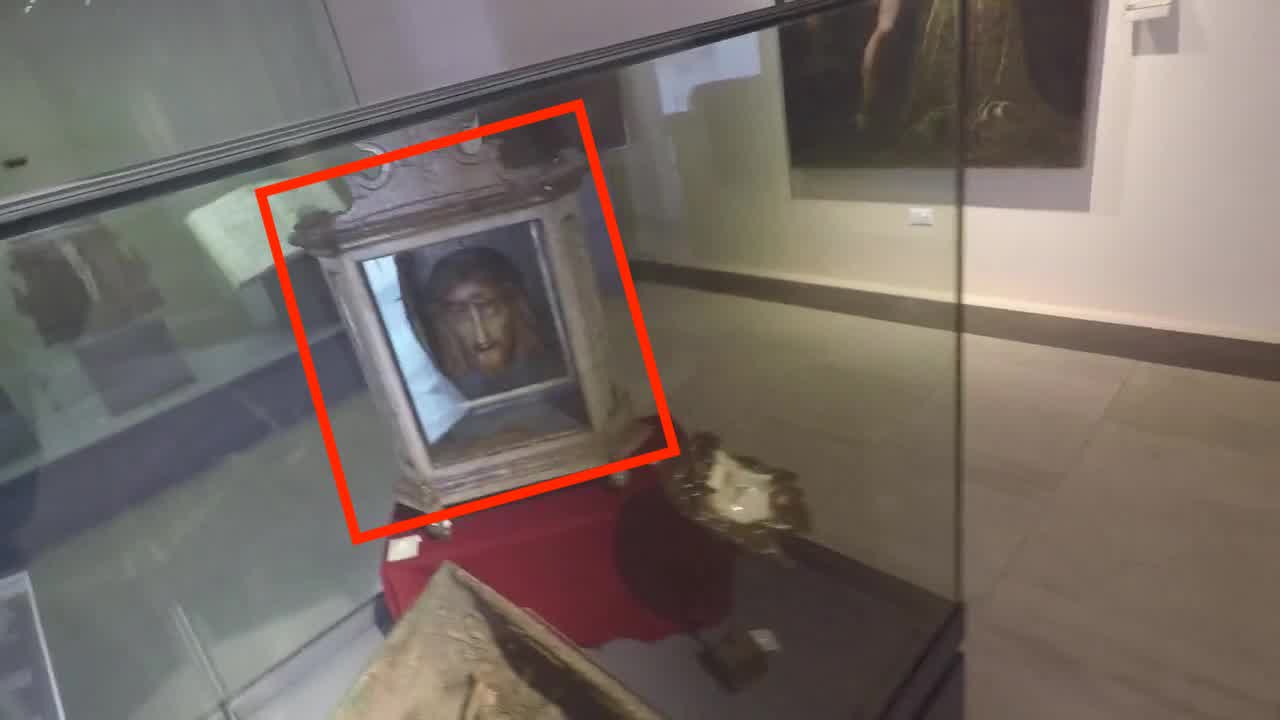}
            \subcaption{Images acquired with GoPro.}
            \caption*{Figure 2: Example of labeled images of the 16 artworks with respect to the three considered domains: (a) Synthetic images, (b) images acquired with HoloLens, (c) images acquired with GoPro.}
            \label{fig:dataset}
\end{figure*}
\begin{table}[]
\caption{Statistics of the proposed dataset for unsupervised multi-target domain adaptation. The average occupied area (last column) is the average percentage of the image occupied by the bounding boxes of the considered object class.}
\includegraphics[width=\linewidth]{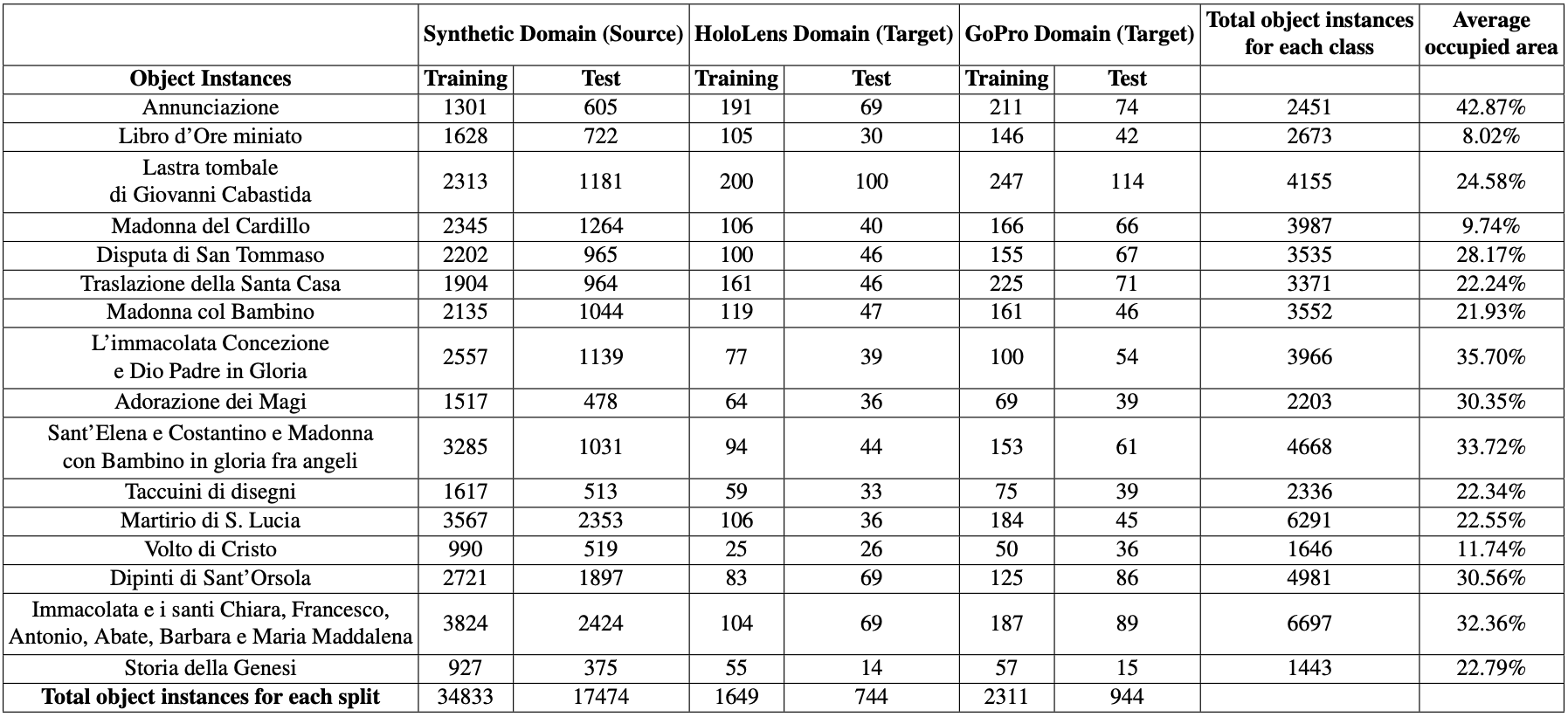}
\label{tab:distribution}
\end{table}

\section{Dataset}
\label{dataset}
To study the problem, we created a dataset\footnote{The dataset is available at \url{https://iplab.dmi.unict.it/OBJ-MDA}} that contains images of 16 artworks included in the cultural site “Galleria Regionale di Palazzo Bellomo\footnote{\url{http://www.regione.sicilia.it/beniculturali/palazzobellomo/}}”. The collection covers different types of artworks, as well as books, sculptures and paintings. We considered three domains: i) synthetic images generated from a 3D model of the cultural site and automatically labeled during the generation process, ii) real images collected by 10 visitors with a HoloLens device and manually labeled, iii) real images collected by the same visitors with a GoPro and manually labeled. Figure 2 shows some examples of images belonging to the three domains. As can be noted, synthetic images differ from the real images in style, shapes of the 3D objects (e.g., observe the statues of Figure 2) and field of view. Similarly, real images acquired with two the different devices differ only in style and field of view.  The three sets of images have been collected as detailed in the following:
\begin{itemize}
    \item Synthetic labeled images (used as source domain): these images have been generated using the tool proposed by~\cite{orlando2020egocentric}. The tool allows to annotate in 3D the position of artworks in the 3D model of a cultural site and simulates an agent navigating the environment while acquiring egocentric images of the observed artworks. The acquired images are automatically labeled by projecting the 3D bounding boxes of the objects onto the generated 2D images. This set contains 75244 images divided in 51284 training images  and 23960 test images.
    \item Target images acquired using a HoloLens: this set of data has been sampled from the work of~\cite{Ragusa_2020} where data has been manually annotated drawing a bounding box around each of the 16 object to match the same artworks present in the synthetic set. This set contains 2190 images divided in 1502 for the training and 688 for the test;
    \item Target images acquired using a Gopro: the dataset was created similarly to the previous one HoloLens. The images have been collected by the same visitor which have visited the site wearing both HoloLens and GoPro wearable cameras. This set contains a total of 2707 images splitted into 1911 for the training and 796 for the test.
\end{itemize}
Table~\ref{tab:distribution} shows the distribution of the object instances in the proposed dataset. As can be noted, the HoloLens and GoPro domains have a number of object instances less than ten times smaller than the synthetic domain. The table also highlights that the proposed dataset is challenging for domain adatation for object detection due to the average size of each object. Indeed, the biggest object present in the dataset occupies only the 42.87\% of the images' area while the smallest occupies 8.02\% of the frame.

\section{Methods}
\label{methods}
In this section, we first give a formal definition of the considered problem. We then discuss the compared methods and present the proposed one.

\subsection{Problem Definition}
Let be S = $\{(x_s^n,y_s^n)\}_{n=1}^{N_s}$ the set of $N_s$ labeled images related to the source domain where $x_s^n$ indicates the $n^{th}$ source image and $y_s^n$ the corresponding annotation. Let $T=\{T_1,T_2,...,T_D\}$ be the set of targets domains where  ${T_i=\{x_{T_i}^n\}_{n=1}^{D_{T_i}}}$ corresponds to the $T_i^{th}$ target domain. We set $D_{T_i}=2$ in our experiments. The goal of unsupervised multi-camera domain adaptation for object detection is to maximize the mAP of the object detector across all the target domains training only on labeled images of the source domain and unlabeled images of the target domains.
\subsection{Baselines without domain adaptation}
We analyze the behaviour of two state-of-the-art object detectors: RetinaNet~\cite{DBLP:journals/corr/abs-1708-02002} and Faster RCNN~\cite{DBLP:journals/corr/RenHG015}. We train and test both detectors on the target domains to produce ``Oracle results" and assess the performance drop observed when the algorithms are trained on synthetic images and tested on the real images of the target domains.
\subsection{Domain adaptation based on feature alignment}
State-of-the-art domain adaptation methods for object detectors commonly consider only one source domain and one target domain. To study whether these state-of-the-art methods can be used to tackle multi-camera domain adaptation, we consider a naive approach which merges the two target domains into a single one. In particular, we considered the following unsupervised domain adaptation methods for object detection: DA-Faster RCNN~\cite{chen2018domain}, Strong Weak~\cite{Saito_2019}, DA-RetinaNet~\cite{PASQUALINO2021104098} and CDSSL~\cite{yu2021unsupervised}.

\subsection{Domain adaptation through feature alignment and image to image translation}
Feature alignment methods aim to reduce the difference between source and target domains at the feature level without taking into account the difference at pixel level (like style, color, shape etc.) which are present between the source and targets domains. For this reason, we combine feature alignment methods with image to image translation methods to reduce the gap also at the pixel level. For the image to image translation task we used the CycleGAN alorithm~\cite{CycleGAN2017} to translate synthetic images to real.

\subsection{Proposed Method}
\begin{figure}[t!]
    \centering
    \includegraphics[width=1\linewidth]{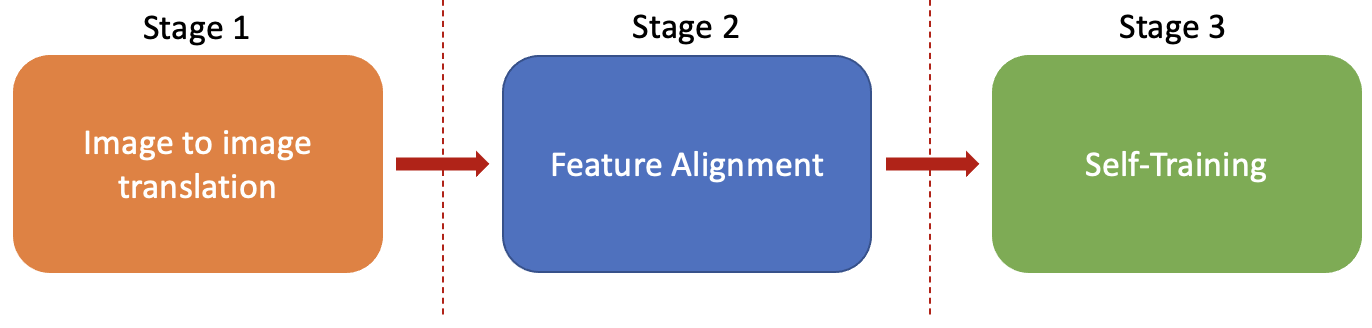}
    \caption{Pipeline of the proposed method. In the Stage 1 synthetic domains is translated to the real domains to reduce the gap at pixel level. In the Stage 2 the gap at feature level is reduced using feature alignment method. In Stage 3 an iterative self-training procedure is used to produce pseudo labels for the target domains.}
    \label{fig:scheme}
\end{figure}
The training of the proposed method comprises three stages that will be discussed in order of execution in the following sections. Each of them contributes to improving the performance of the object detector and works to adapt the two distribution at different levels. Figure~\ref{fig:scheme} shows an overview of the general pipeline of the proposed method.
\subsubsection{Image to Image Translation}
\begin{figure}[t!]
    \centering
    \includegraphics[width=1\linewidth]{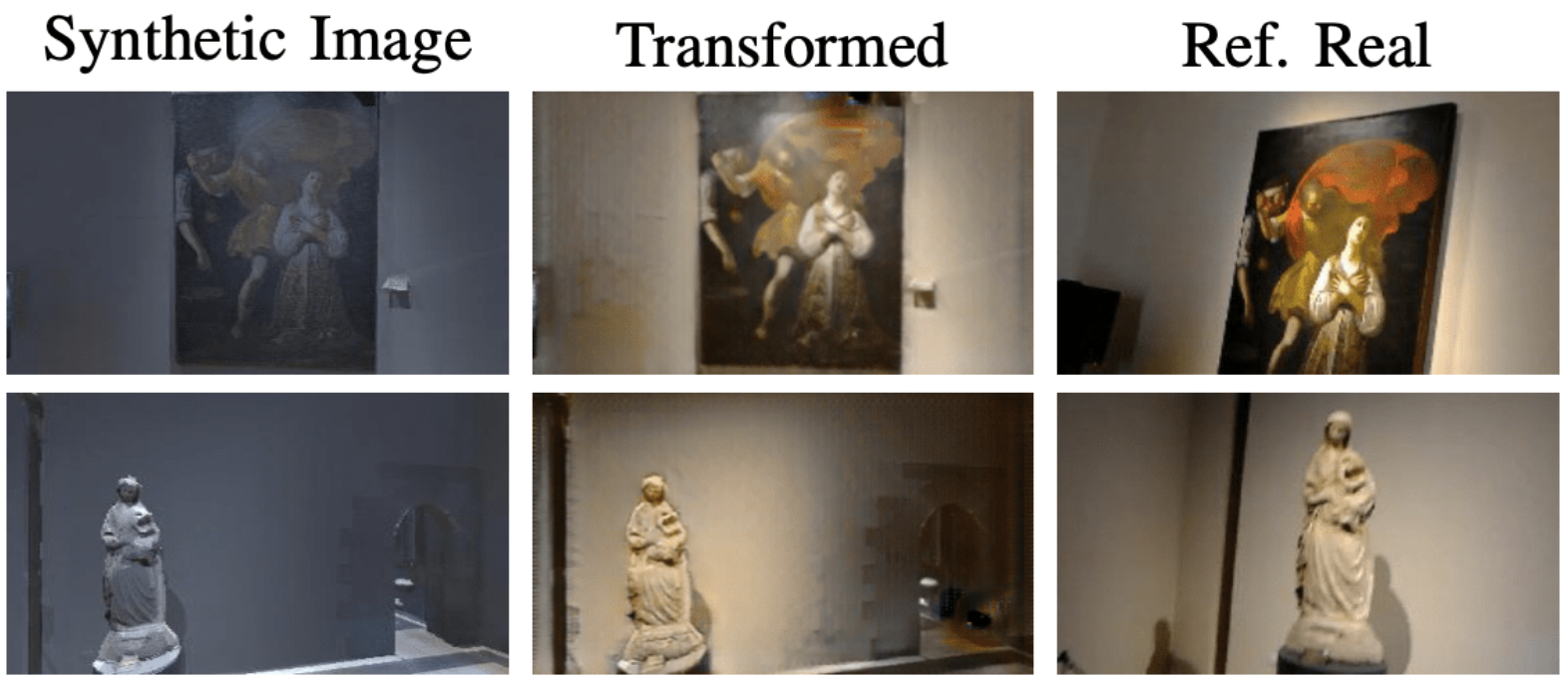}
    \caption{Qualitative results obtained using CycleGAN as image to image translation method (Stage 1). Synthetic images (left) are translated to the merged target domains (center). Real images similar to the translated ones are also reported for reference (right).}
    \label{fig:Synthetic_to_real}
\end{figure}
Synthetic images generated from a model acquired using a 3D scanner, such as Matterport~\footnote{\url{https://matterport.com/}}, differ in general from real images in the style and shape of the object which can affect object detection performance. To reduce this diversity, the first step of our method consists in mitigating the style and shape differences using an image to image translation method. In particular, we used CycleGAN to transform training synthetic images into the real. In the later stages of our pipeline, the object detection model will be trained on the transformed images and tested directly on the real images. This step is optional in our pipeline for two reasons: 1) it can be computationally expensive when the datasets are large; 2) when the target and the source domain are not too similar, this transformation can be not sufficiently accurate. Figure~\ref{fig:Synthetic_to_real} shows some qualitative results of this translation. As can be noted, the transformed images look more similar to the real counterpart after the transformation.

\subsubsection{Feature Alignment}
\label{feature_al_sub}
\begin{figure}[t!]
    \centering
	\includegraphics[width=1\linewidth]{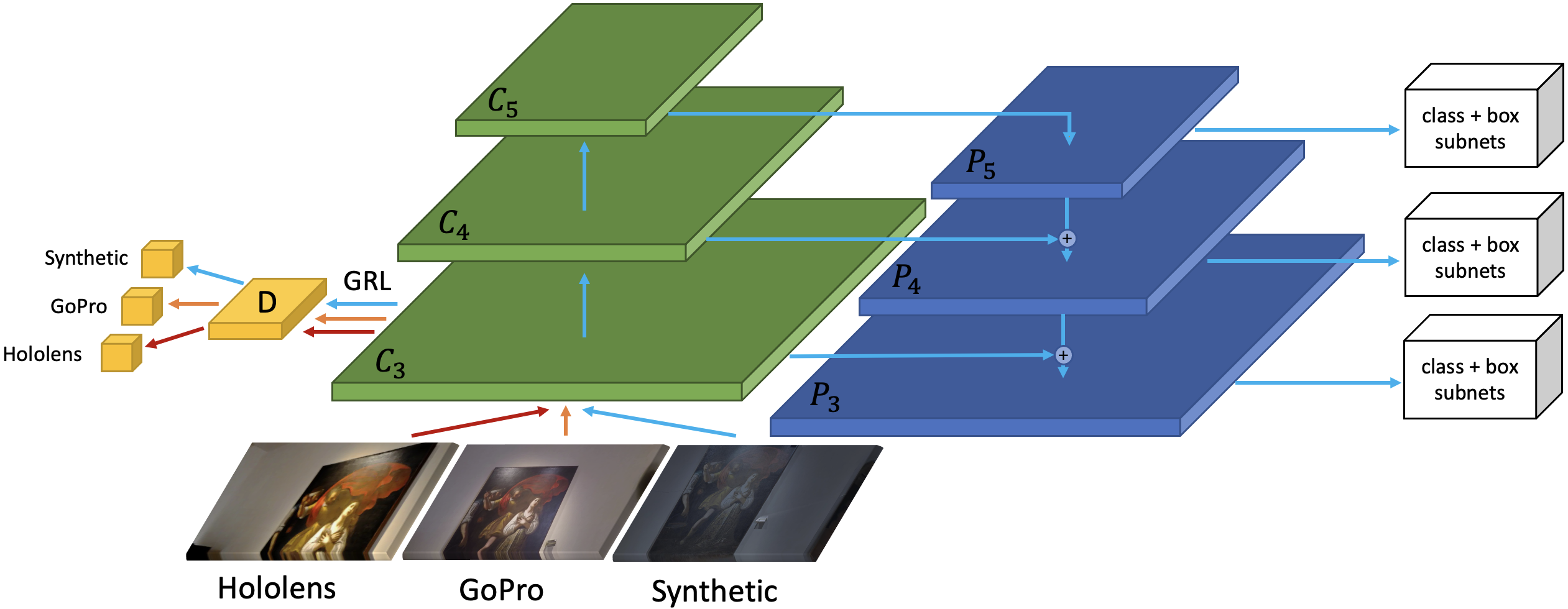} 
	\caption{Architecture of the proposed MDA-RetinaNet model.}
	\label{fig:mda-s1} 
\end{figure}
Although image to image translation can be used to reduce the differences in terms of style and shape, the features extracted from the two domains can still be different. For this reason, in this second stage we propose an object detection architecture which jointly adapts the features during the training. \newline
State-of-the-art domain adaptation methods for object detection do not consider the existence of multiple target domains. To take advantage of multiple unlabeled target domains during the training, we propose a model, that we called MDA-RetinaNet, to address the problem of unsupervised domain adaptation for object detection based on adversarial learning~\cite{ganin2014unsupervised}. Figure~\ref{fig:mda-s1} shows the architecture of the proposed method which builds on RetinaNet~\cite{DBLP:journals/corr/abs-1708-02002}. To reduce the domain gap present at the feature level, we attach a domain discriminator with a gradient reversal layer to the feature map $C_3$ obtained from the ResNet backbone~\cite{he2016deep}. In particular, to adapt multiple domains (in this case 1 source and 2 targets in our experiments) we consider a multi-class classifier $D$ which discriminates among all of them.
The discriminator has 3 convolutional layers with kernel size equal to 1, followed by a ReLU activation function. Following~\cite{ganin2014unsupervised}, we place a gradient reversal layer at the input of the discriminator and train the model by minimizing the following loss function:
\begin{center} 
    $L = L_{class} + L_{box} - \lambda(L_D)$
\end{center}
where $L_{class}$ and $L_{box}$ are the regression losses of RetinaNet, $L_D$ is the loss of the discriminator module and $\lambda$ is an hyper-parameter that balances the object detection and domain adaptation losses. This approach differs from standard methods that use a binary classifier used to only discriminate features belonging to the source and target domains, hence ignoring the presence of multiple targets. We hypothesize that, providing a multi-classes discriminator, the model will learn to extract features which are not only indistinguishable across synthetic and real domains, but also indistinguishable across the different real cameras. It is important to highlight that this type of adaptation allow to learn a combination of weights that extract feature maps using the backbone that generalize to the different domains. No adaptation is directly enforced for the layers involved in the classification and regression of the bounding boxes (Figure~\ref{fig:mda-s1} white modules).
\subsubsection{Self-Training}
\label{selfTraining}
 \begin{figure}[t]
    \centering
	\includegraphics[width=1\linewidth]{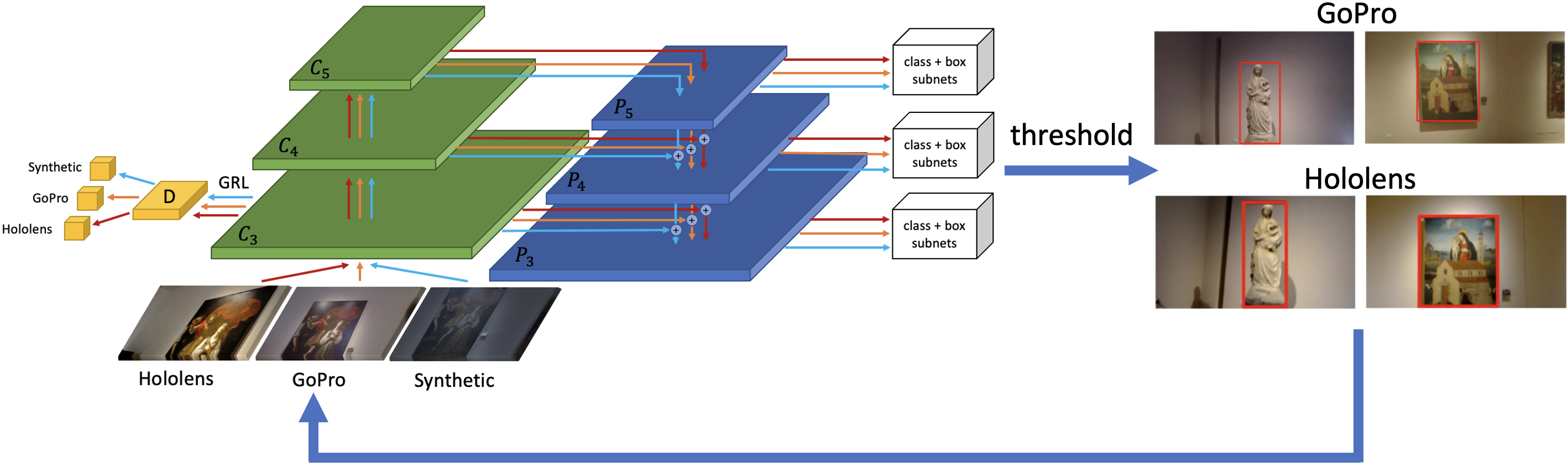} 
	\caption{Self-training module for MDA-RetinaNet.}
	\label{fig:mda-s2} 
\end{figure}
As noted in the previous section~\ref{feature_al_sub}, the adaptation provided in the second stage is at the level of the features extracted by the backbone, whereas the classification and regression layers which detect the objects are trained only with the synthetic images due to the absence of labels for the real images. To tackle this limitation, we generate pseudo labels for the real images by retraining the model with the predictions on real images above a given confidence threshold produced by the model trained at the second stage of our pipeline (see Figure~\ref{fig:scheme}). This allows to train MDA-RetinaNet in a supervised way as illustrated in Figure~\ref{fig:mda-s2} by exploiting the obtained pseudo labels. This latest module is trained in an iterative way, gradually increasing the threshold used to generate the pseudo labels to reduce the error of potentially wrong predicted labels. Algorithm~1 reports the complete procedure of the proposed method.
\begin{algorithm}[t!]
\SetAlgoLined
 \textbf{Input:} $S = \{x_s^n,y_s^n\}$ the source domain, $T= \{T_1,T_2,...T_D\}$ target domains, $converge = false$;\\
 \textbf{Step 1:} transform the set $S$ into $T$ using CycleGAN;\\
 \textbf{Step 2:} train MDA-RetinaNet using $S'$ and $T$ to adapt the features;\\
 \textbf{Step 3:} set the threshold $t = 0.75$ and produce the pseudo labels $y^n_d$ for each images in each target $T_1,T_2,..,T_D$ using MDA-RetinaNet;\\
 \textbf{Step 4:}\\ 
 \While{!converge}{
  train MDA-RetinaNet using $y_d^n$\;
  produce the new pseudo label $y_d^n$\;
  \eIf{$t < 0.9$}{
    $t = t + 0.05$\;
   }{
   $converge = true;$
   }
 }
 \textbf{Output:} B - the set of predicted bounding boxes.
 \caption{Proposed multi-target domain adaptation for object detection algorithm.}
\end{algorithm}
\subsection{Experimental Settings}
All the compared models were trained for 60K iterations using weights pretrained on ImageNet~\cite{deng2009imagenet}. We set the learning rate to 0.0002 for the first 30K iterations, then we multiply it by 0.1 for the remains 30K iterations. DA-Faster RCNN, Strong Weak and CDSSL were trained with the same parameters proposed by the authors in their respective papers~\cite{chen2018domain,Saito_2019,yu2021unsupervised}.
The batch size was set to 4 for RetinaNet, 6 for DA-RetinaNet (4 source and 2 target images, 1 HoloLens and 1 GoPro) and 8 for MDA-RetinaNet\footnote{code available at https://github.com/fpv-iplab/STMDA-RetinaNet} (4 source and 4 target images divided in 2 HoloLens and 2 GoPro images). MDA-RetinaNet was implemented using Detectron2~\cite{wu2019detectron2}. To reduce the noise of the initial training of the Discriminator D, we adapt the $\lambda$ hyperparameter following the update rule proposed in~\cite{ganin2014unsupervised}. The second stage (Section~\ref{feature_al_sub}) is performed only one time to produce initial pseudo labels. The Self-Training stage (Section~\ref{selfTraining}) is executed 4 times gradually increasing the threshold used to generate the new pseudo labels which will be used in the next iteration. 
CycleGAN was trained for 60 epochs using the default parameters.

\section{Results}
\label{results}
This Section reports and analyzes the results of the experimental analysis.
\subsection{Feature Alignment Results}
Table~\ref{tab:feature_alignment} reports the results of the feature alignment based models. The first two rows show the results of the baseline Faster RCNN and RetinaNet modules trained with synthetic images and tested on HoloLens and GoPro without any domain adaptation technique. It is worth noting that RetinaNet is less sensitive to the domain gap, obtaining an mAP $\sim$7\% higher than Faster RCNN (14.10\% vs 7.61\% on HoloLens and 30.39\% vs 37.13\% on GoPro). For this reason, we focused our further experiments considering RetinaNet as backbone for the object detector in our proposed methods. The second group of rows (rows 3-6 of Table~\ref{tab:feature_alignment}) report the results of state-of-the-art methods adapted for this specific task. In particular, due to the fact that these methods are able to work only with a single target, we merged the HoloLens and GoPro datasets into one. The proposed MDA-RetinaNet performs better than the other models and outperforms the best state-of-the-art method, DA-RetinaNet, by $\sim$3\% for HoloLens (34.97\% vs 31.63\%) and $\sim$2\% for GoPro (50.81\% vs 48.37\%). The last row shows the results of MDA-RetinaNet combined with the self-training procedure. As the results highlight, this combination allows to increase the performances of $\sim$23\% if compared with DA-RetinaNet (54.36\% vs 31.63\%) for HoloLens, $\sim$11\% (59.51\% vs 48.37\%) for GoPro and $\sim$20\% if compared with MDA-RetinaNet without self-training (54.36\% vs 34.94\%) for HoloLens and $\sim$9\% (59.51\% vs 50.81\%) for GoPro. Furthermore, the performance gap between HoloLens and GoPro with this last model it is almost negligible.
\begin{table}[t]
\centering
\caption{Results of baseline and feature alignment methods. S refers to Synthetic, H refers to HoloLens and G to GoPro. ST indicates the self-training procedure.}
\label{tab:feature_alignment}
\centering
\begin{tabular}{|c|c|c|c|c|c|}
\hline
Model & Source & Target & Test H & Test G \\
\hline
Faster RCNN \cite{DBLP:journals/corr/RenHG015} & S & - & 7.61\% & 30.39\% \\
\hline
RetinaNet \cite{DBLP:journals/corr/abs-1708-02002} & S & - & 14.10\% & 37.13\% \\
\hline
\hline
DA-Faster RCNN \cite{chen2018domain} & S & H+G & 10.53\% & 48.23\% \\
\hline
Strong Weak \cite{Saito_2019} & S & H+G &  26.68\% & 48.55\% \\
\hline
CDSSL~\cite{yu2021unsupervised} &  S & H+G &  28.66\% & 45.33\%\\
\hline
DA-RetinaNet \cite{PASQUALINO2021104098} & S & H+G & 31.63\% & 48.37\% \\
\hline
MDA-RetinaNet & S &  H, G & 34.97\% & 50.81\% \\
\hline
MDA-RetinaNet + ST & S &  H, G & \textbf{54.36\%} & \textbf{59.51\%} \\
\hline
\hline
Faster RCNN \cite{DBLP:journals/corr/RenHG015} (Oracle) & H &- & 91.97\% & 76.88\% \\
\hline
Faster RCNN\cite{DBLP:journals/corr/RenHG015} (Oracle) & G & - & 68.65\% &89.21\% \\
\hline
RetinaNet~\cite{DBLP:journals/corr/abs-1708-02002} (Oracle) & H & - & 92.44\% & 77.96\% \\
\hline
\textcolor{black}{RetinaNet~\citep{DBLP:journals/corr/abs-1708-02002} (Oracle)} & \textcolor{black}{G} & \textcolor{black}{-} & \textcolor{black}{69.70\%} & \textcolor{black}{89.69\%} \\
\hline
\end{tabular}
\end{table}

\subsection{Feature Alignment and Image to Image translation Results}
Table~\ref{tab:cycle_feature} shows the results obtained combining the baseline and feature alignment methods with CycleGAN. The first two rows report the results of Faster RCNN and RetinaNet when trained on synthetic images transformed to the merged HoloLens and GoPro domain. As can be noted, pixel level domain adaptation allows to significantly increase the performance of Faster RCNN and RetinaNet respectively by about 8\% (7.61\% vs 15.34\%) and  16\% (14.10\% vs 31.43\%) on HoloLens and by about 33\% (30.39\% vs 63.60\%) and 32\% (37.13\% vs 69.59\%) on GoPro, reducing the gap between synthetic and real images.
The middle part of the table shows the results of the methods based on feature alignment. Also in this case, MDA-RetinaNet achieves an higher mAP with respect to the best state-of-the-art method, CDSSL, (53.06\% vs 58.11\% for HoloLens and 71.17\% vs 71.39\% for GoPro) which further improves if we introduce the self-training procedure (58.11\% vs 66.64\% for HoloLens and 71.39\% vs 72.22\% for GoPro). It is worth noting that, with self-training the gap in performances between HoloLens and GoPro is reduced from $\sim$13\% to $\sim$6\% which suggest that the model acquires knowledge from the GoPro images that is useful to detect object in the HoloLens domain.
Furthermore, the performance of MDA-RetinaNet with self-training is really close to the performance of the RetineNet oracles when trained with the labeled HoloLens domain and tested on GoPro and vice versa (66.64\% vs 69.70\% for HoloLens and 72.22\% vs 77.96\% for GoPro). However, there is still space of improvement if we consider the performances of the oracles trained and tested in their respective domains, which makes proposed dataset still challenging (66.64\% vs 92.44\% for HoloLens and 72.22\% vs 89.69\% for GoPro).
\subsection{Ablation Study}
Table~\ref{tab:ablation4} reports the ablation study of the proposed MDA-RetinaNet model and compares the results with respect to the DA-RetinaNet architecture. We evaluated the models on HoloLens domain, which is more challenging if compared to GoPro, analyzing the impact of the placement of the discriminator at different levels of the feature map extracted from the RetinaNet backbone (see Figure~\ref{fig:mda-s1} on the paper). As can be noted, each single discriminator increases the performances of the standard RetinaNet architecture and obtain better performances than DA-RetinaNet. The discriminator attached to the first feature map $C_3$, allows to achieve better results than the other two discriminators attached to the $C_4$ and $C_5$ feature maps. Moreover, considering more than one discriminator to align the feature at different levels does lead to obtain better results in our experiments as in the case of the single domain DA-RetinaNet but only decreases the performance.
\textcolor{black}{The best combination and optimal number of discriminators was found empirically and, as shown in Table~\ref{tab:ablation4}, it is achieved using only one discriminator at the $C_3$ level. We hypothesize that considering more discriminators at the same time could unbalance the models training, obtaining features that are aligned but less effective for the main object detection task.}
\textcolor{black}{In Table~\ref{tab:ablation4} we also report an ablation study of the impact of each discriminator attached at $P_i$ or at $C_i$ levels. As can be noted, in each case, the performances achieved by the models that use the discriminator at $P_i$ levels are lower than their counterparts which use discriminators at the $C_i$ levels.
Table~\ref{tab:ablation5} shows the results obtained with different linear schedules of the values of the threshold. We noted that, due to the domain gap between source and target domains, it is convenient to use a low threshold in the first iterations of self-training, where a set of initial pseudo-labels is needed, and increasing this threshold to an higher value as training proceeds. Indeed, we achieve best results for using a threshold value starting at 0.75 and ending at 0.9.
Table~\ref{tab:ablation3} reports the results of adapting DA-Faster RCNN~\cite{chen2018domain} and Strong Weak~\cite{Saito_2019} to multiple target domains using the same methodology proposed for MDA-RetinaNet. Specifically, instead of merging the to dataset into one and use the binary discriminator proposed by the authors in their papers, we replaced it with our multi classes discriminator and considered the target domains individually instead of merging them. As can be noted, the performances of the other two methods improves by ~3-4\% if compared with the results of Table~\ref{tab:feature_alignment}. Nevertheless, the best results are still obtained by the proposed MDA-RetinaNet architecture. These results suggest that using a multi class discriminator instead of a binary discriminator allows to consistently improve performances with different architectures.}
\begin{table}[t]
\centering
\caption{Results of feature alignment methods combined with CycleGAN. H refers to HoloLens while G to GoPro. ``\{G,~H\}" refers to synthetic images translated to the merged HoloLens and GoPro domains. ST indicates self-training procedure.}
\label{tab:cycle_feature}
\centering
\begin{tabular}{|c|c|c|c|c|c|}
\hline
Model & Source & Target & Test H & Test G \\
\hline
Faster RCNN  \cite{DBLP:journals/corr/RenHG015} & \{G, H\} & - & 15.34\% & 63.60\% \\
\hline
RetinaNet~\cite{DBLP:journals/corr/abs-1708-02002} & \{G, H\} & - & 31.43\% & 69.59\% \\
\hline
\hline
DA-Faster RCNN \cite{chen2018domain} & \{G, H\} & H+G & 32.13\% & 65.19\% \\
\hline
Strong Weak~\cite{Saito_2019}  & \{G, H\} & H+G & 41.11\% & 66.45\% \\
\hline
DA-RetinaNet  \cite{PASQUALINO2021104098}  & \{G, H\} & H+G & 52.07\% & 71.14\% \\
\hline
CDSSL~\cite{yu2021unsupervised}  & \{G, H\} & H+G & 53.06\% & 71.17\% \\
\hline
MDA-RetinaNet  & \{G, H\} & H, G & \textbf{58.11}\% & \textbf{71.39\%}\\
\hline
MDA-RetinaNet + ST  & \{G, H\} & H, G & \textbf{66.64}\% & \textbf{72.22\%}\\
\hline
\hline
Faster RCNN \cite{DBLP:journals/corr/RenHG015} (Oracle) & H & - & 91.97\% & 76.88\% \\
\hline
Faster RCNN \cite{DBLP:journals/corr/RenHG015} (Oracle) & G & - & 68.65\% & 89.21\% \\
\hline
RetinaNet~\citep{DBLP:journals/corr/abs-1708-02002} (Oracle) & H & - & 92.44\% & 77.96\% \\
\hline
RetinaNet~\citep{DBLP:journals/corr/abs-1708-02002} (Oracle) & G & - & 69.70\% & 89.69\% \\
\hline
\end{tabular}
\end{table}

\begin{table}[h!]
\caption{\textcolor{black}{Ablation study about the impact of each discriminator $D_i$ and comparison between each discriminator $D_i$ placed at $C_i$ and $P_i$ level.}}
\label{tab:ablation4}
\centering
\textcolor{black}{
\begin{tabular}{|c||c|c||c|c||c|c||c|}
\hline
Model & $C_3$& $P_3$ & $C_4$ & $P_4$ & $C_5$ & $P_5$ & mAP\\
\hline
RetinaNet &  &  &  & & & & 14.10\%\\
\hline
DA-RetinaNet &  &  & & &\checkmark & & 15.84\%\\
\hline
MDA-RetinaNet &  & & & & \checkmark & & 19.54\%\\
\hline
MDA-RetinaNet &  &  & & & & \checkmark & 16.29\%\\
\hline
DA-RetinaNet &  & &\checkmark & & & & 16.38\%\\
\hline
MDA-RetinaNet &  & & \checkmark & & & & 19.88\%\\
\hline
MDA-RetinaNet &  & & &\checkmark & & & 17.01\%\\
\hline
DA-RetinaNet & \checkmark & & & &  &  & 28.61\%\\
\hline
MDA-RetinaNet & \checkmark & & & & & & \textbf{34.97\%}\\
\hline
MDA-RetinaNet & & \checkmark & & & &  & 31.44\%\\
\hline
DA-RetinaNet & \checkmark & &\checkmark &  & & &30.52\%\\
\hline
MDA-RetinaNet & \checkmark & &\checkmark & & & & 34.09\%\\
\hline
MDA-RetinaNet & &\checkmark & &\checkmark & & & 30.85\%\\
\hline
DA-RetinaNet & \checkmark & & \checkmark & & \checkmark & & 31.04\%\\
\hline
MDA-RetinaNet & \checkmark & & \checkmark & & \checkmark & & 32.11\%\\
\hline
MDA-RetinaNet & & \checkmark & & \checkmark & & \checkmark & 30.18\%\\
\hline
\end{tabular}
}
\end{table}

\begin{table}[h!]
\centering
\caption{\textcolor{black}{Comparison performance considering different threshold.}}
\label{tab:ablation5}
\textcolor{black}{
\begin{tabular}{|c|c|c|c|c|}
\hline
Model  & Threshold & Test H & Test G \\
\hline
MDA-RetinaNet + ST & 0.90 & 47.48\% & 52.25\% \\
\hline
MDA-RetinaNet + ST & 0.85 to 90 & 49.21\% & 54.90\% \\
\hline
MDA-RetinaNet + ST &  0.80 to 0.90 & 52.49\% & 57.67\% \\
\hline
MDA-RetinaNet + ST & 0.75 to 0.90 & \textbf{54.36\%} & \textbf{59.51\%} \\
\hline
\end{tabular}
}
\end{table}

\begin{table}[t]
\centering
\footnotesize
\caption{\textcolor{black}{Comparison between DA-Faster RCNN, Strong Weak and MDA-RetinaNet when modified using multiclass discriminators. S refers to Synthetic, H refers to Hololens and G to GoPro.}}
\textcolor{black}{
\begin{tabular}{|c|c|c|c|c|c|}
\hline
Model & Source & Target & Test H & Test G \\
\hline
DA-Faster RCNN \cite{chen2018domain} & S & H, G & 13.79\% & 48.35\% \\
\hline
Strong Weak \cite{Saito_2019} & S & H, G & 29.52\% & 49.06\% \\
\hline
MDA-RetinaNet & S &  H, G & \textbf{34.97\%} & \textbf{50.81\%} \\
\hline
\end{tabular}
}
\label{tab:ablation3}
\end{table}

\begin{table}[t]
\centering
\footnotesize
\caption{Comparison between DA-RetinaNet trained using one target set at a time and MDA-RetinaNet. S refers to Synthetic, H refers to Hololens and G to GoPro.}
\begin{tabular}{|c|c|c|c|c|c|}
\hline
Model & Source & Target & Test H & Test G \\
\hline
RetinaNet~\cite{DBLP:journals/corr/abs-1708-02002} & S & - & 14.10\% & 37.13\% \\
\hline
DA-RetinaNet \cite{PASQUALINO2021104098} & S & H & 31.01\% & 36.60\% \\
\hline
DA-RetinaNet \cite{PASQUALINO2021104098} & S & G & 21.63\% & 45.86\% \\
\hline
MDA-RetinaNet & S &  H, G & \textbf{34.97\%} & \textbf{50.81\%} \\
\hline
\end{tabular}
\label{tab:ablation2}
\end{table}
\subsection{Comparison between MDA-RetinaNet and DA-RetinaNet}
Table~\ref{tab:ablation2} compares the results of the proposed MDA-RetinaNet with DA-RetinaNet. It is worth noting that training the model using only one target domain at a time results in worse performance in both domains despite they are very similar. This happens because the model overfits with respect to the considered target domain used for training. Using both domains during training, as the proposed MDA-RetinaNet model does, allows to generalize over both target domains with a single model, which also results in improved performance.

\subsection{Qualitative Results}
\begin{figure}[t!]
            \centering
            \begin{minipage}{.3\textwidth}
            \centering
             \small{RetinaNet}\\
            \end{minipage}
            \begin{minipage}{.3\textwidth}
            \centering
             \small{MDA-RetinaNet}\\
            \end{minipage}
            \begin{minipage}{.3\textwidth}
            \centering
             \small{MDA-RetinaNet+ST}\\
            \end{minipage}
            
            \vspace{1mm}
            \includegraphics[width=.3\textwidth]{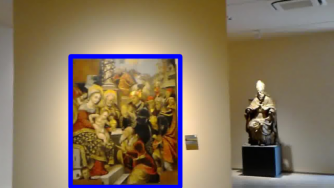}
            \includegraphics[width=.3\textwidth]{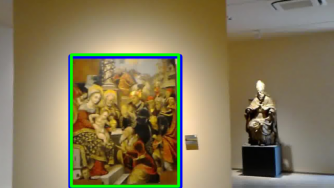}
            \includegraphics[width=.3\textwidth]{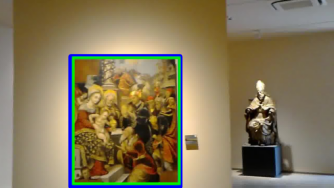}

            \vspace{1mm}
            \includegraphics[width=.3\textwidth]{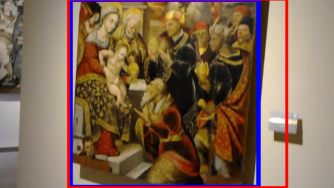}
            \includegraphics[width=.3\textwidth]{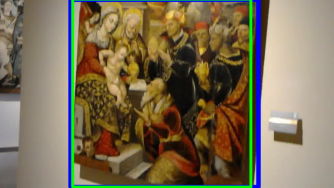}
            \includegraphics[width=.3\textwidth]{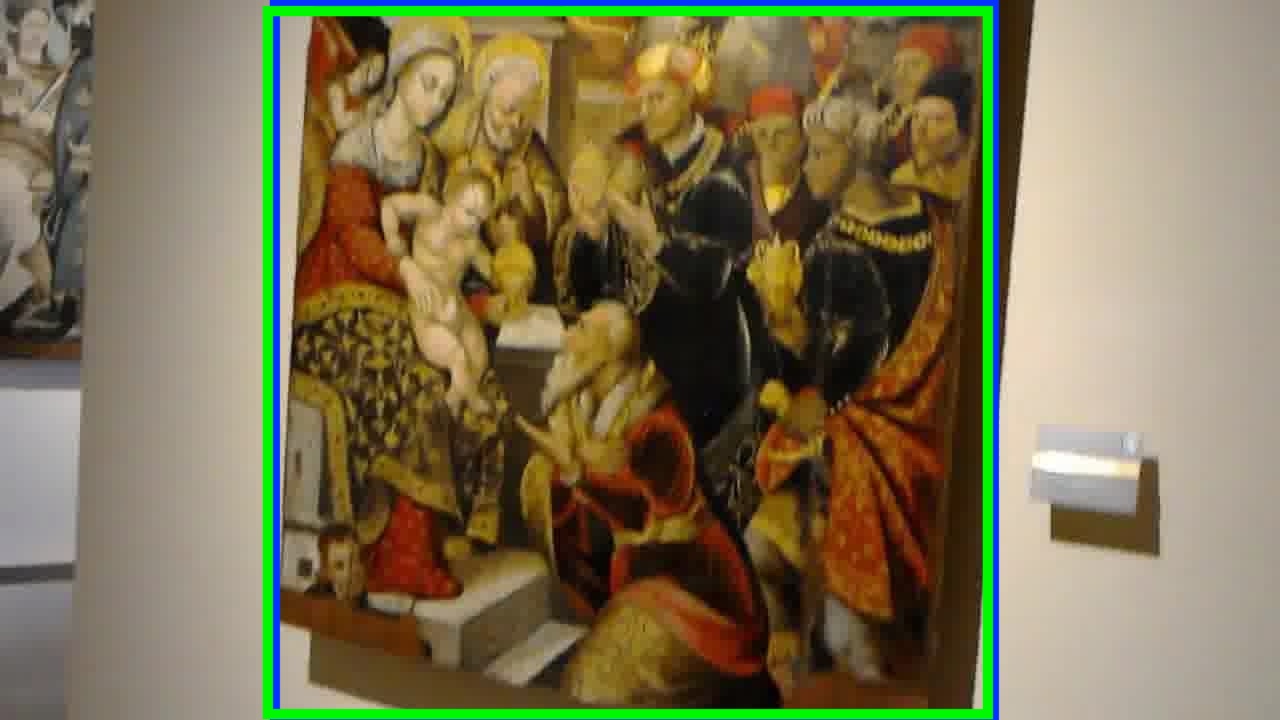}

            \vspace{1mm}
            \includegraphics[width=.3\textwidth]{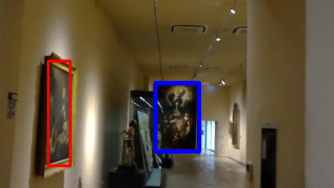}
            \includegraphics[width=.3\textwidth]{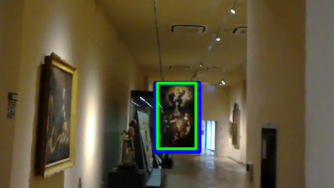}
           \includegraphics[width=.3\textwidth]{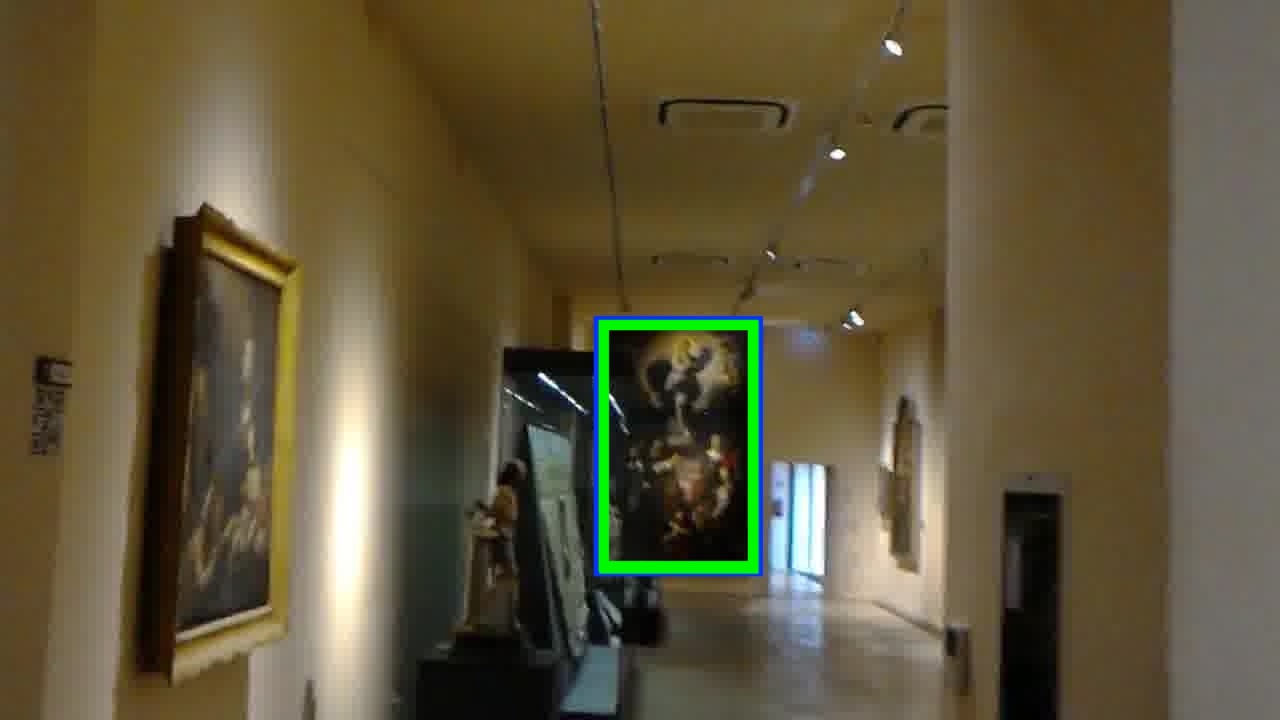}

            \vspace{1mm}
            \includegraphics[width=.3\textwidth]{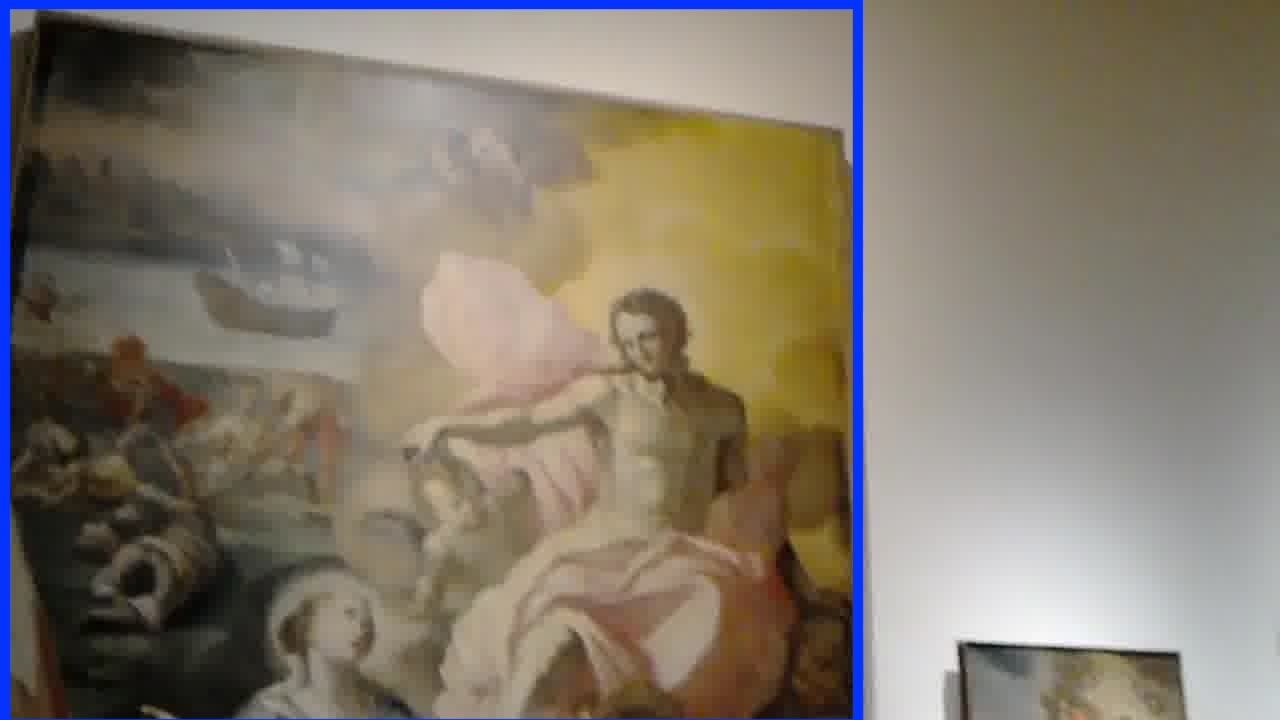}
            \includegraphics[width=.3\textwidth]{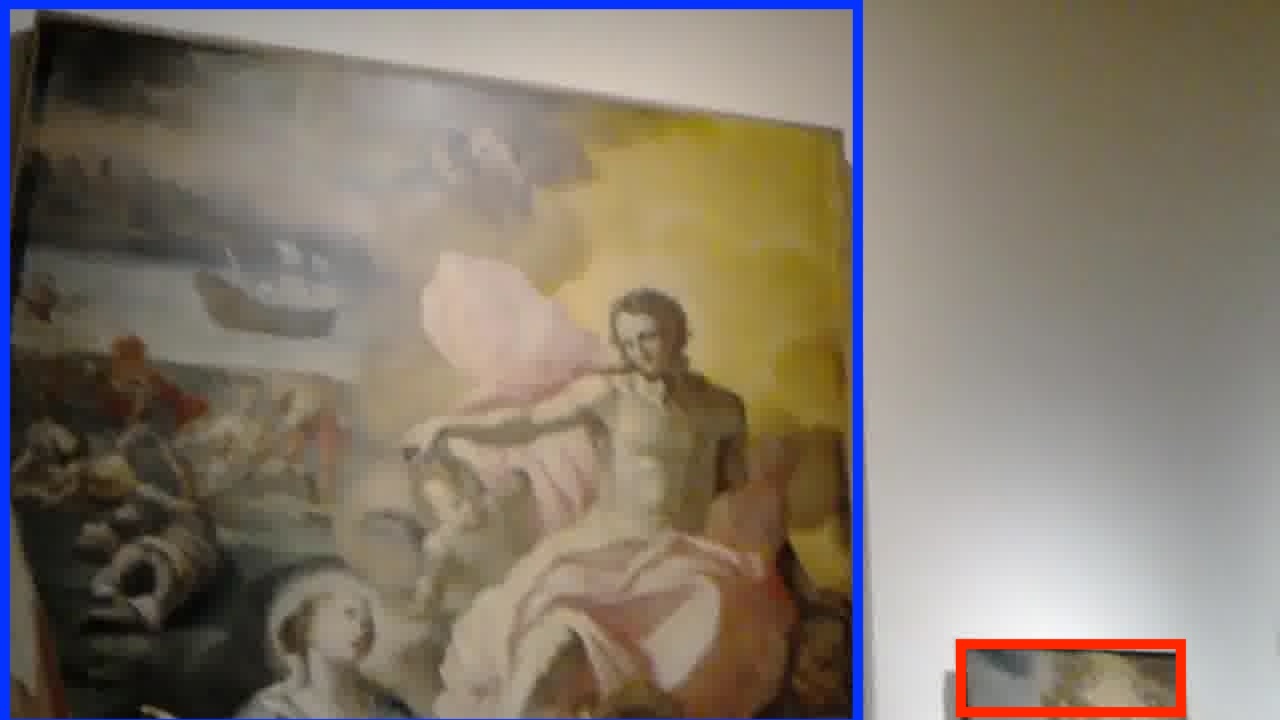}
            \includegraphics[width=.3\textwidth]{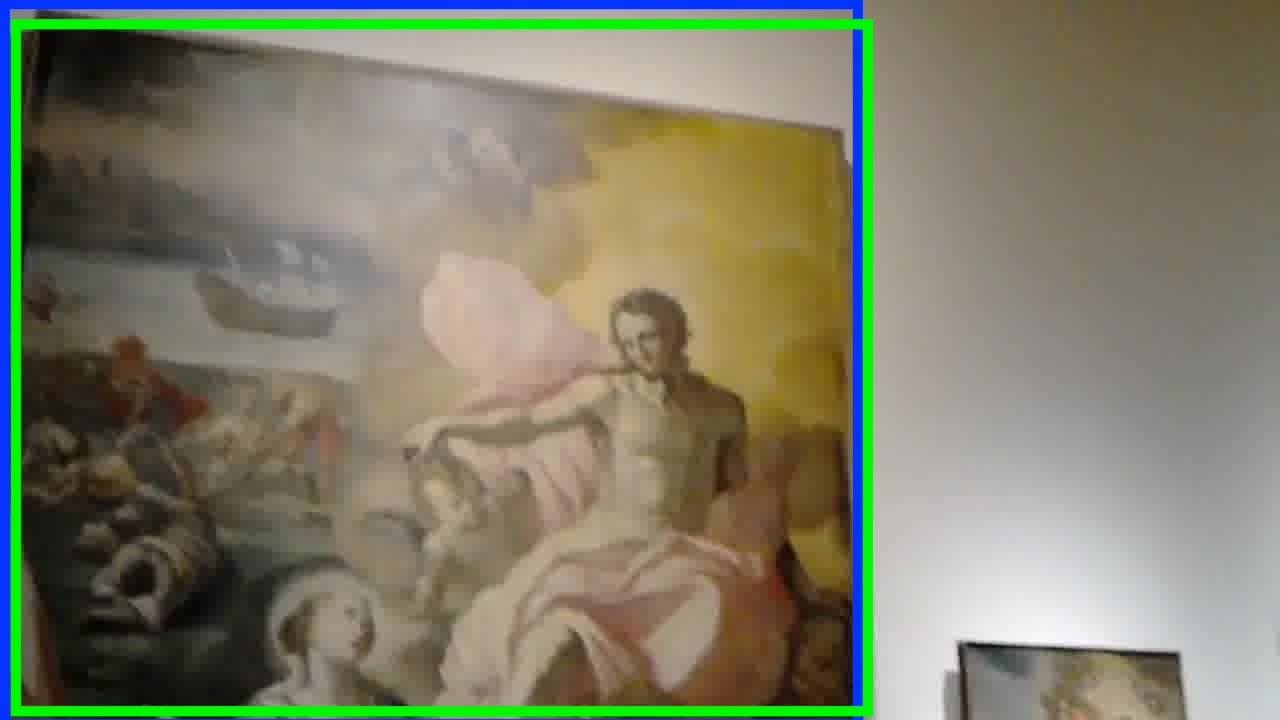}
            
            \vspace{1mm}
            \includegraphics[width=.3\textwidth]{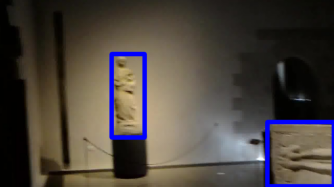}
            \includegraphics[width=.3\textwidth]{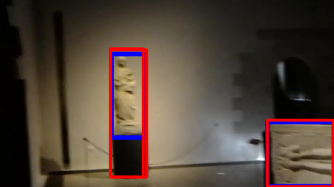}
            \includegraphics[width=.3\textwidth]{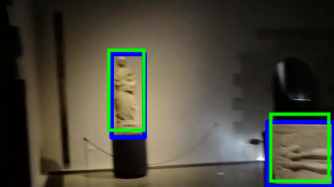}
            \caption{Qualitative results of RetinaNet, MDA-RetinaNet and MDA-RetinaNet with self-training (ST). The blue box represents ground truth, the red box indicates a wrong detection (object localization or classification), the green box represents correct detections.}
            \label{fig:qualitativeresult}
\end{figure}
Figure~\ref{fig:qualitativeresult} compares some qualitative detection results obtained by the proposed MDA-RetinaNet with and without Self-Training with respect to RetinaNet baseline (the ground truth is the blue bounding box). RetinaNet fails the detection in many cases. Indeed, it does not detects any artwork or produce a wrong classification and/or regression. MDA-RetinaNet well recognize small and large artworks but fails in the last two rows. MDA-RetinaNet with Self-Training improve the performance of the standard RetinaNet and MDA-RetinaNet with a more accurate detection of the artworks.

\section{Conclusion}
\label{conclusion}
We studied the problem of unsupervised multi-camera domain adaptation for object detection in cultural sites. To perform the study, we have collected and publicly released a new challenging dataset with the aim to encourage the community to continue researching on the problem. We proposed a new method which combines feature alignment, pixel level and self-training methods that outperforms current state-of-the-art methods. 
\section*{Acknowledgments}
This research has been supported by the project VALUE
(N. 08CT6209090207 - CUP G69J18001060007) - PO FESR 2014/2020 - Azione 1.1.5., by the project MEGABIT - Research Program  Pia.ce.ri. 2020/2022 Linea 2 - University of Catania, and by Project HERO - Huge and Easy Reproduction of Objects.


\bibliography{mybibfile}

\end{document}